\documentclass[times,twocolumn,final,longtitle]{elsarticle}

\usepackage{marvosym}
\usepackage{eso-pic}

\newcommand{\arxiv}{1} 

\ifdefined\arxiv
    \usepackage{medima-arxiv}
    \newcommand{\writer}[1]{~}
    \newcommand{\orcid}[1]{\href{https://orcid.org/#1}{\includegraphics[scale=0.09]{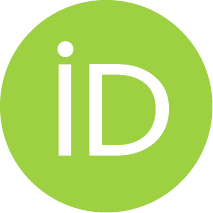}}}
    \newcommand{\mailto}[1]{\href{mailto:#1}\textsuperscript{\textsuperscript{\Letter,}} }
    \journal{}
    \date{}

\else
    \usepackage{medima}
    \usepackage{dblfloatfix} 

\makeatletter
\@ifpackageloaded{endfloat}{%
  \let\efloat@iwrite\relax
  
  \let\@elt\relax
}{}
\makeatother
    \newcommand{\writer}[1]{}
    \newcommand{\orcid}[1]{}
    \newcommand{\mailto}[1]{}
\fi

\usepackage{framed}
\usepackage{makecell}
\usepackage{multirow}
\usepackage{multicol}
\usepackage{tabularx}
\usepackage{booktabs}
\usepackage{amssymb}
\usepackage{amsmath}
\usepackage{latexsym}
\usepackage[obeyspaces]{url}
\usepackage[dvipsnames]{xcolor}
\usepackage{rotating}
\usepackage{makecell}
\usepackage{pifont}
\usepackage{array}
\usepackage{etoolbox}

\definecolor{newtext1}{RGB}{0,0,0}
\definecolor{newtext}{RGB}{0,0,0}

\usepackage[colorlinks=true,citecolor=blue,urlcolor=blue]{hyperref}
\AtBeginDocument{%
  \hypersetup{
    citecolor=teal,
    linkcolor=blue,   
    urlcolor=Blue}}

\newcolumntype{P}[1]{>{\centering\arraybackslash}p{#1}}
\newcommand{\teal}{\color{teal}}

\newcommand{\team}[3]{%
    \subsubsection{{\bf #1:~}%
    {\noindent\it\teal#2}}%
    {\noindent#3}
}

\preto\tabular{\setcounter{magicrownumbers}{0}}
\newcounter{magicrownumbers}

\definecolor{newcolor}{rgb}{.8,.349,.1}
\journal{Medical Image Analysis}

\begin{document}
\AddToShipoutPictureFG*{%
  \AtTextUpperLeft{%
    \put(0,24){%
      \makebox(\textwidth,0)[c]{%
        \footnotesize\itshape
        Accepted for publication in Medical Image Analysis%
      }%
    }%
  }%
}
\verso{Alapatt \textit{et~al.}}

\begin{frontmatter}

\title{{\bf The SAGES Critical View of Safety Challenge: A Global Benchmark for AI-Assisted Surgical Quality Assessment}}%


\author[Unistra,Scialytics]{
Deepak \snm{Alapatt}\corref{cor1}\fnref{fn1}}\ead{deepak@scialytics.io}
\author[MGH,Cologne]{Jennifer \snm{Eckhoff} \fnref{fn1}}
\author[MGH]{Zhiliang \snm{Lyu}}
\author[MGH,SJTU]{Yutong \snm{Ban}}
\author[IHU]{Jean-Paul \snm{Mazellier}}
\author[Northwell,Albany]{Sarah \snm{Choksi}}
\author[SJTU]{Kunyi \snm{Yang}}

\author[IHU,ChanGung]{Po-Hsing \snm{Chiang}}
\author[UBologna,IRCAD``]{Noemi \snm{Zorzetti}}
\author[upen]{Samuele \snm{Cannas}}

\author[theator]{Daniel \snm{Neimark}}
\author[theator]{Omri \snm{Bar}}
\author[DKFZ]{Amine \snm{Yamlahi}}
\author[DKFZ]{Jakob \snm{Hennighausen}}
\author[Stanford]{Xiaohan \snm{Wang}}
\author[Stanford]{Rui \snm{Li}}
\author[WithAI]{Long \snm{Liang}}
\author[WithAI]{Yuxian \snm{Wang}}
\author[NAAMII]{Saurabh \snm{Koju}}
\author[NAAMII,aberdeen]{Binod \snm{Bhattarai}}
\author[Eindhoven]{Tim \snm{Jaspers}}
\author[UCL]{Zhehua \snm{Mao}}
\author[UCL]{Anjana \snm{Wijekoon}}
\author[UHN]{Jun \snm{Ma}}
\author[Transformers]{Yinan \snm{Xu}}
\author[Cologne]{Zhilong \snm{Weng}}
\author[URV]{Ammar M. \snm{Okran} \orcid{0000-0001-9264-0301}}
\author[URV]{Hatem A. \snm{Rashwan} \orcid{0000-0001-5421-1637}}
\author[HUST]{Boyang \snm{Shen}}
\author[HUST]{Kaixiang \snm{Yang}}
\author[HFUT]{Yutao \snm{Zhang}}
\author[HFUT]{Hao \snm{Wang}}
\author[]{SAGES CVS Challenge Data Consortium}

\author[MGH]{Quanzheng \snm{Li}}
\author[Northwell]{Filippo \snm{Filicori}}
\author[MGH]{Xiang \snm{Li}}
\author[IHU,gemelli]{Pietro \snm{Mascagni}}
\author[MGH,upen,duke]{Daniel A. \snm{Hashimoto}}
\author[duke,MGH]{Guy \snm{Rosman} \orcid{0000-0002-9334-1706}}
\author[MGH,duke]{Ozanan \snm{Meireles} \fnref{fn2}}
\author[Unistra,IHU]{Nicolas \snm{Padoy} \fnref{fn2}}

\address[Unistra]{University of Strasbourg, CNRS, INSERM, ICube, UMR7357, Strasbourg, France}
\address[IHU]{IHU Strasbourg, Strasbourg, France}
\address[Scialytics]{Scialytics SAS, France}
\address[MGH]{Massachusetts General Hospital, Harvard Medical School, USA}
\address[Cologne]{University Hospital Cologne, Germany}
\address[SJTU]{Global College, Shanghai Jiao Tong University, China}
\address[ChanGung]{Chang Gung Memorial Hospital, Keelung, Taiwan}
\address[UBologna]{A. Costa Hospital and Maggiore Hospital - AUSL Bologna, Bologna, Italy}
\address[IRCAD]{Institute for Research against Digestive Cancer (IRCAD), Strasbourg, France}
\address[Northwell]{Lenox Hill Hospital, Northwell Health, USA}
\address[Albany]{Albany Medical Center, USA}

\address[theator]{theator, Israel}
\address[DKFZ]{German Cancer Research Center (DKFZ), Germany}
\address[Stanford]{Stanford University, USA}
\address[WithAI]{ChengDu Withai Innovations Technology Company, China}
\address[NAAMII]{MultiModal Learning Lab, NAAMII, Nepal}
\address[aberdeen]{University of Aberdeen, UK}
\address[Eindhoven]{Eindhoven University of Technology, Netherlands}
\address[UCL]{UCL Hawkes Institute, Dept of Computer Science, University College London, London, UK}
\address[UHN]{University Health Network, Canada}
\address[Transformers]{Institute for Biomedical Informatics, Germany}
\address[URV]{Rovira i Virgili University, Tarragona, Spain}
\address[HUST]{Huazhong University of Science and Technology, China}
\address[HFUT]{Hefei University of Technology, China}

\address[gemelli]{Fondazione Policlinico Universitario A. Gemelli IRCCS, Italy}
\address[upen]{University of Pennsylvania, USA}
\address[duke]{Duke University, USA}

\cortext[cor1]{Corresponding author.}

\fntext[fn1]{Deepak Alapatt and Jennifer Eckhoff contributed equally and share co-first authorship.}
\fntext[fn2]{Ozanan Meireles and Nicolas Padoy contributed equally and share co-last authorship.}
\fntext[fn3]{The SAGES CVS Challenge Data Consortium is listed in the Acknowledgment section.}


\begin{abstract}
Advances in artificial intelligence (AI) for surgical quality assessment promise to democratize access to expertise, with applications in training, guidance, and accreditation. This study presents the SAGES Critical View of Safety (CVS) Challenge, the first AI competition organized by a surgical society, using the CVS in laparoscopic cholecystectomy, a universally recommended yet inconsistently performed safety step, as an exemplar of surgical quality assessment. A global collaboration across 54 institutions in 24 countries engaged hundreds of clinicians and engineers to curate 1,000 videos annotated by 20 surgical experts according to a consensus-validated protocol. The challenge addressed key barriers to real-world deployment in surgery, including achieving high performance, capturing uncertainty in subjective assessment, and ensuring robustness to clinical variability. To enable this scale of effort, we developed EndoGlacier, a framework for managing large, heterogeneous surgical video and multi-annotator workflows. Thirteen international teams participated, achieving up to a 17\% relative gain in assessment performance, over \textcolor{newtext}{62}\% reduction in calibration error, and a 17\% relative improvement in robustness over the state-of-the-art. Analysis of results highlighted methodological trends linked to model performance, providing guidance for future research toward robust, clinically deployable AI for surgical quality assessment.
\end{abstract}

\begin{keyword}
\vspace{-0.33in}
\KWD Surgical activity recognition\sep Critical View of Safety \sep Cholecystectomy \sep Safety \sep SAGES \sep Challenge.
\end{keyword}

\end{frontmatter}



\section{Introduction}

\begin{figure*}[ht]
\center
    \includegraphics[width=0.9\linewidth]{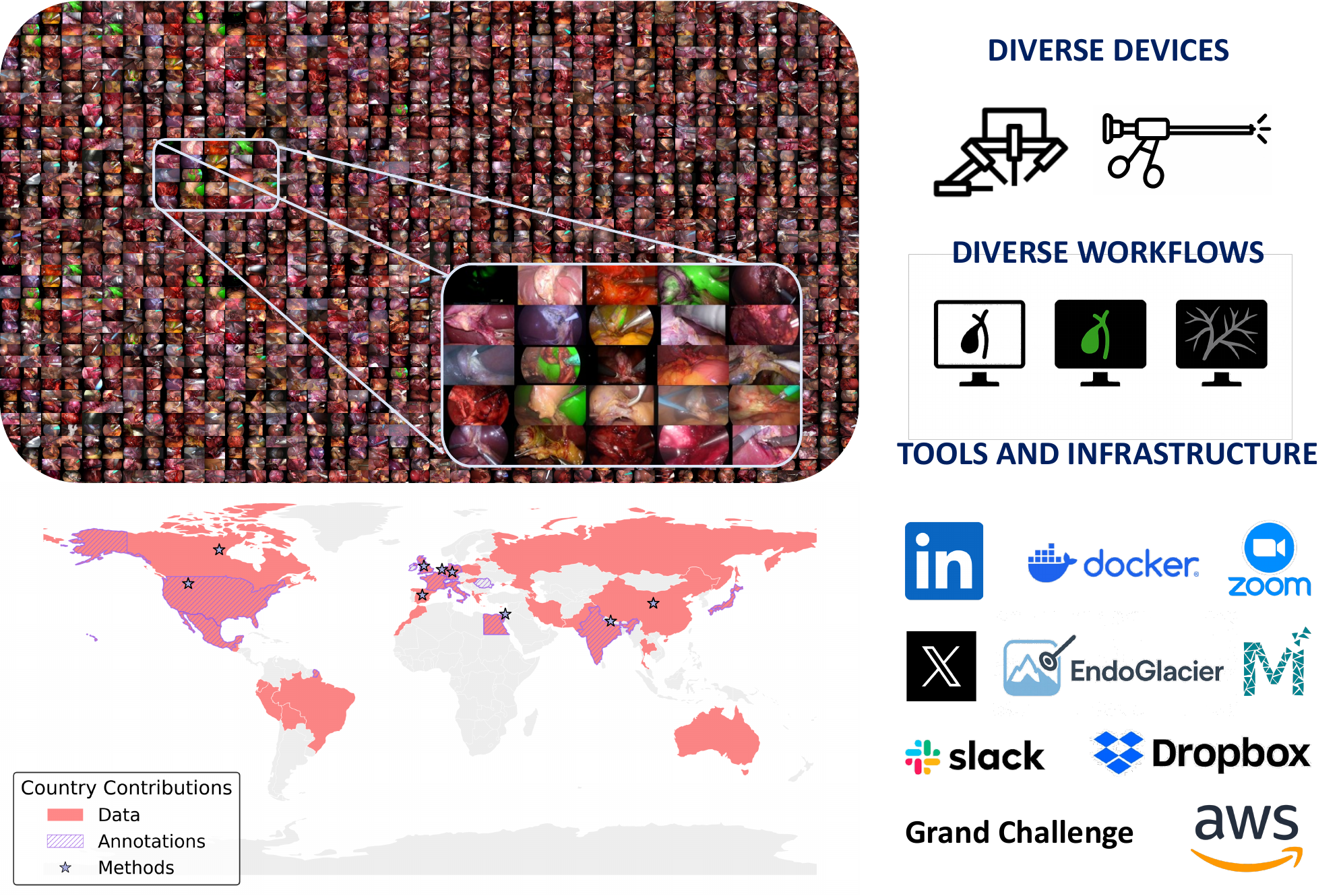} 
    \caption{A global benchmark for surgical safety AI, the challenge brought together over 1000 annotated cholecystectomy videos, representing diverse acquisition devices (hardware, instrumentation, etc.) and workflows (case difficulty, techniques, preferences, instrumentation, etc.). The dataset was constructed through a global collaborative effort spanning three years, with the bottom-left panel illustrating countries contributing data, annotations, and methods to this international initiative.}
    \label{fig:challenge_snapshot}
\end{figure*}

Surgery is among the most critical components of modern healthcare, accounting for about a third of all healthcare expenditure and a comparable share of the global burden of disease \citep{meara2015global}. Despite this awesome financial and human impact, surgical practice remains variable, both in processes and outcomes. Historically, the operating room (OR) has functioned as a data-poor environment, limiting scalable efforts to understand and reduce this variability. The digital transformation of the OR, particularly driven by the widespread adoption of minimally invasive surgery, has begun to shift this trend. Surgical video, now a routine byproduct of surgical care, is enabling a new paradigm in which data-driven approaches can uncover the link between the quality of surgical care and patient outcomes.

Surgical quality, spanning decision making, technical skills, and adherence to best practices, is now a central focus in understanding how intraoperative factors influence patient outcomes. Since the seminal work of \cite{birkmeyer2013surgical} demonstrating a quantifiable link between technical skills and patient outcomes, video based assessment has emerged as a powerful tool for evaluating surgical quality. Surgical societies, most notably the Society of American Gastrointestinal and Endoscopic Surgeons (SAGES)\footnote{Society of American Gastrointestinal and Endoscopic Surgeons (SAGES). \url{https://www.sages.org}},
have led efforts in this space through dedicated initiatives aimed at advancing the science of video-based surgical assessment. However, despite growing interest, most existing efforts are constrained by limited scale. The experts best equipped to assess surgical quality, experienced surgeons, are often too burdened to evaluate large volumes of video. Compounding this issue is the inherent subjectivity in surgical assessment, with variability even among expert reviewers. Artificial intelligence offers a natural solution to these challenges, with the potential to replicate and disseminate expert consensus. These AI assessments could democratize access to expertise when it's needed the most, in the OR, and enable a mature understanding of the variability that we have come to accept as an intrinsic part of surgical care today.

A notable example of progress toward AI-assisted surgical quality assessment is the automated evaluation of the Critical View of Safety (CVS) \citep{strasberg1995analysis} in laparoscopic cholecystectomy. CVS is a universally recommended safety step that ensures adequate dissection for the safe identification of anatomical structures, yet it remains inconsistently implemented in clinical practice. Prior research has demonstrated the feasibility of using AI to assess CVS achievement \citep{mascagni2022artificial}, document this step \citep{mascagni2021computer}, and even provide real-time intraoperative feedback \citep{mascagni2024early}. However, most existing approaches rely on black-box deep learning models, raising important concerns about their robustness. Variations in surgical instrumentation, technique, workflow, lighting conditions, and image quality across centers introduce significant challenges to generalization—challenges that must be rigorously understood before these models can safely influence clinical decisions.

To thoroughly evaluate and improve AI models for surgical quality assessment, structured biomedical challenges offer a powerful framework. These challenges allow researchers to compete and benchmark diverse methodological approaches on curated datasets. The SAGES CVS Challenge represents a landmark step in this direction. As the first biomedical challenge organized by a surgical society, it leverages the unique infrastructure of the Society of American Gastrointestinal and Endoscopic Surgeons (SAGES), whose 7,000+ members include global leaders in minimally invasive surgery. In this challenge, CVS serves as an exemplar of surgical quality: a well-defined, clinically meaningful target with clear guidelines for assessment. Through a global effort with 54 institutions across 24 countries, we assembled a diverse dataset of 1,000 laparoscopic cholecystectomy videos, each curated with rigorous quality-control protocols and annotated by a panel of 20 expert surgeons.

This work describes the resulting benchmark and the EndoGlacier infrastructure that enabled it: an automation-driven framework for coordinating global video sourcing, expert training, multi-annotator workflows, and quality control. Participants were evaluated across three complementary dimensions essential for clinical deployment—absolute performance requirements, alignment with expert uncertainty, and robustness under real-world domain shifts. Submissions from 13 international teams delivered measurable gains over state-of-the-art in all three areas, offering both methodological advances and insight into the design choices that influence performance. By uniting scale, clinical focus, and reproducible evaluation, this benchmark aims to catalyze the development of robust AI systems for surgical quality assessment.

\section{Related work}
The SAGES CVS challenge relates to several research topics, for which we present the relevant literature in the following paragraphs.

\subsection{Video-based Surgical Quality Assessment}
The first work linking technical skills to outcomes dates back to 2013 \citep{birkmeyer2013surgical}, and since then, the growing body of literature on the topic \citep{gruter2023video} has prompted the need for solutions to measure surgical performance objectively. Minimally invasive surgery and surgical video are quickly becoming some of the most promising sources of information to make these assessments. Being able to track how performance changes over time, essentially the learning curve, means we can begin to shift toward competency-based credentialing models \citep{pryor2023american} and offer personalized feedback \citep{naik2018personalized}. Beyond the individual surgeon, this could help identify systematic issues and offer potential solutions at the department, hospital, or even healthcare system level. The applications are innumerable, fundamentally boiling down to being able to identify gaps in current practice and offer actionable insights and support. Most quality assessment tools, scoring rubrics used to measure quality, generally fall into one of four categories: assessing errors (i.e., suboptimal techniques), assessing events (i.e., consequential errors), technical skills (e.g., precision, smoothness), and procedure-specific assessments (e.g., quality of performing a certain step, choice of technique) \citep{gruter2023video}.

\textbf{Relative positioning of our work}:
Previous work on manual video-based surgical quality assessment \citep{gruter2023video} is increasingly validating the potential, if not the need, for systems to perform objective assessments.

\textcolor{newtext1}{Most manual video-assessment studies have been designed to validate an assessment tool or its association with clinical outcomes within a relatively bounded cohort, rather than to establish a broadly sourced benchmark. Of the 52 manual assessment studies catalogued by \citet{gruter2023video}, 49 reported the number of evaluated videos. Forty-eight of these studies (98\%) evaluated fewer than 250 videos, and 42 (86\%) evaluated fewer than 100. The only prior study larger than the present benchmark evaluated 1{,}220 videos through a Japanese national surgical qualification programme \citep{matsuda2014reliability}; the next-largest evaluated 217 videos \citep{crochet2021performance}. Reviewer-pool size was reported for 47 studies, of which 33 (70\%) used five or fewer reviewers and 38 (81\%) used fewer than 20. Among the 46 studies reporting both video-set and reviewer-pool size, only \citet{matsuda2014reliability} combined at least 250 videos with more than five reviewers.}

\textcolor{newtext1}{Our benchmark combines 1{,}000 videos drawn from 54 institutions in 24 countries with a pool of 20 qualified annotators. Each video received three independent assessments, yielding 3{,}000 video-level assessments. Together with \citet{matsuda2014reliability}, it is therefore one of only two manual assessment initiatives in this literature to combine a video set of at least 250 cases with a reviewer pool larger than five, and the only such initiative constructed across multiple countries. This combination of scale, source diversity, and annotation diversity exposes Critical View of Safety assessment to variation in acquisition devices, institutional practices, operative technique, and expert interpretation, providing a substantially more demanding benchmark for methods intended to operate on real-world surgical video.}

\subsection{AI-based Surgical Quality Assessment}
Alongside the manual video-based assessment literature, there is a growing body of work demonstrating the feasibility and value of automating surgical quality assessments. To date, much of the work in this space has focused on technical skill evaluation \citep{lam2022machine}, driven in part by the lower data privacy concerns when working with simulator data and the availability of rigorously validated scoring frameworks. More recently, efforts have begun to extend to other types of quality metrics, such as the detection of adverse events \citep{bose2025feature,eppler2023automated} like bleeding, thermal injuries, or mechanical damage to anatomical structures. Procedure-specific models are also becoming more common, with several focused on evaluating the adequacy of safety step implementation \citep{mascagni2022artificial} or the approach taken during key procedural moments \citep{laplante2023validation}. Downstream applications of these models are increasingly being demonstrated, from automated documentation of safety steps \citep{mascagni2021computer} to evaluation of training interventions \citep{mascagni2022artificial}, and even delivery of real-time intraoperative feedback \citep{mascagni2024early} and predictive analytics~\citep{yin2024hypergraph}.

\textbf{Relative positioning of our work}: Our work falls within the category of procedure-specific assessments of surgical quality, and extends it by foregrounding several understudied but critical considerations for real-world deployment. We explore and benchmark methods that explicitly account for the inherent ambiguity in surgical quality assessments. We also systematically evaluate the robustness of these methods under distribution shifts related to device characteristics, case complexity, regional differences in technique, and other real-world factors. In doing so, we contribute a valuable benchmark and highlight key failure modes that warrant deeper investigation if these models are to be used safely and reliably in clinical settings.

\subsection{AI for CVS assessment}
Among the various initiatives in surgical quality, Critical View of Safety (CVS) assessment stands out for its well-established clinical relevance and the growing volume of work exploring its automation. Initially proposed by \cite{strasberg1995analysis}, CVS is a safety step designed to ensure adequate dissection for the secure identification of anatomical landmarks, thereby preventing bile duct injuries—a devastating complication that costs over a billion dollars annually in the United States alone \citep{carroll1998common}. Now supported by more than 30 years of literature, CVS is universally recommended by multi-society guidelines \citep{brunt2020safe} and is known to prevent over 97\% of major bile duct injuries when properly implemented \citep{way2003causes}. Despite strong evidence suggesting that CVS can be achieved in 90–95\% of cases \citep{avgerinos2009one,sanjay2010critical,tsalis2015open}, its real-world implementation rate remains alarmingly low, often between 10–15\% \citep{korndorffer2020situating}. This discrepancy has been attributed to the subjectivity of the assessment, as well as overconfidence among surgeons.

In response to this critical gap, researchers have developed structured annotation protocols \citep{mascagni2021surgical}, AI models for CVS assessment and documentation \citep{korndorffer2020situating,mascagni2022artificial,mascagni2021computer,murali2023latent,murali2023encoding,ban2023concept,yin2024hypergraph}, and systems to provide intraoperative feedback \citep{mascagni2024early}. Over time, a growing body of literature has emerged around this task, including multiple public benchmarks \citep{rios2023cholec80,murali2023endoscapes,mascagni2025endoscapes} and purpose-built methodological frameworks.

\textbf{Relative positioning of our work}: We build on this strong foundation by validating, refining, and expanding previously proposed annotation protocols through structured expert consensus, and by assembling the largest and most diverse benchmark dataset in this domain. Not only does our dataset exceed prior efforts by an order of magnitude in scale, with 1,000 laparoscopic cholecystectomy videos, but it also includes annotations from a more diverse panel of expert reviewers (20 annotators). The data spans a wide range of acquisition settings, including different institutions, devices, and regional practices, with representation from low- and middle-income countries. This diversity enables new lines of investigation into sources of bias related to annotators, devices, geography, and workflow, pushing the field toward more robust and generalizable AI models for surgical safety.

\subsection{Biomedical challenges}
Biomedical challenges are a well-established format prompting teams from academia and industry alike to compete in advancing research or benchmarking on emerging or underserved topics. Within surgical computer vision, challenges have regularly served as a springboard for insight into promising research directions, especially in establishing new benchmarks in this relatively niche domain \citep{maier2022surgical}. To date, most endoscopic vision challenges have focused on foundational tasks aimed at understanding surgical context from video data. These efforts span a range of spatial and temporal granularities—from coarse procedural phases \citep{maier2021heidelberg} to fine-grained actions \citep{nwoye2023cholectriplet2021,nwoye2023cholectriplet2022}, from frame-level assessments \citep{al2019cataracts} to pixel-precise segmentations—across various types of surgical procedures \citep{allan20192017,allan20202018,ross2021comparative,maier2021heidelberg,luengo20212020,bano2021fetreg}.

\textbf{Relative positioning of our work}: Leveraging the platform of SAGES, and as the first biomedical challenge led by a surgical society, our work seeks to bridge two persistent gaps in the challenge literature: clinical relevance and scale. Building on prior challenges that have successfully advanced foundational, context-aware systems for surgical video understanding, we pivot upstream toward a task with a clearly defined and direct clinical purpose: the automatic assessment of the Critical View of Safety. In doing so, we surface critical issues that emerge when moving from procedural understanding to clinical quality assessment, such as managing subjectivity in labeling, capturing ambiguity in human assessment, and building models robust to real-world data heterogeneity.

In terms of scale, our challenge departs from the common approach of sourcing data and annotations from one or a small number of institutions. Prior challenges have demonstrated immense value, but their limited scale can introduce subtle biases tied to local practices, individual annotators, or specific devices. By expanding the dataset across 54 centers in 24 countries and involving a diverse panel of globally dispersed annotators, we open up new possibilities for studying generalization, robustness, and fairness in clinical AI models. This scale also brings previously hidden challenges to the surface, ranging from how to design evaluation protocols that account for inter-rater variability, to the technical and logistical hurdles of coordinating global annotation efforts.

Naturally, scaling up biomedical challenges in this way introduces new demands. From data governance and privacy compliance to the infrastructure needed to engage, track, and manage a large number of stakeholders, we encountered and addressed a range of practical challenges. We hope that the \textcolor{newtext1}{workflow design and implementation lessons from} this challenge serve as a framework for future initiatives operating at similar or larger scales, especially those seeking to tackle clinically meaningful tasks with real-world constraints.
\section{Coordinated Data Flows for Large-Scale Annotation}
\label{sec:endoglacier}

\begin{figure*}[t]
    \includegraphics[width=0.945\linewidth]{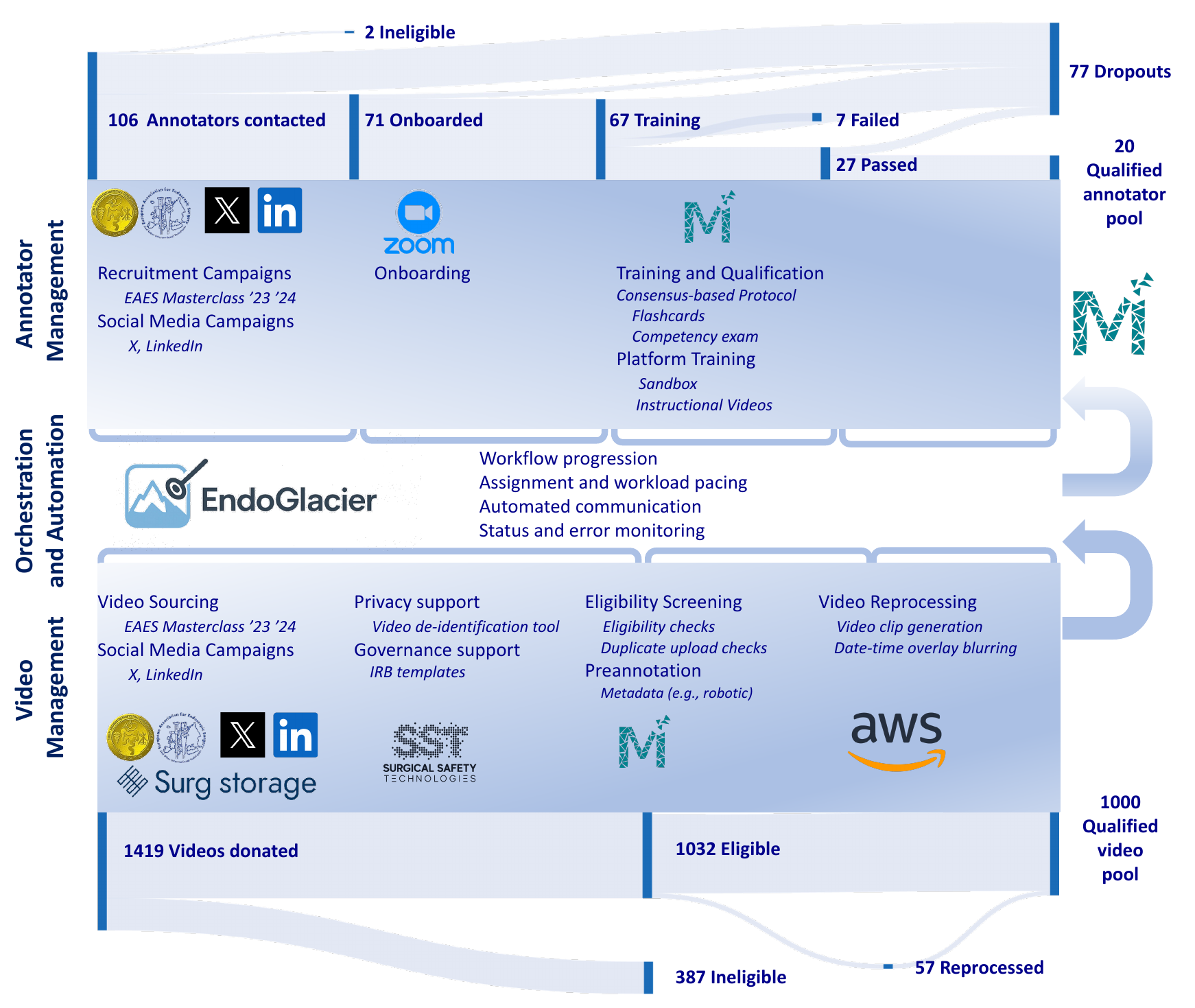} 
    \caption{Overview of the annotation infrastructure. The pipeline integrated three coordinated flows: (i) Annotator management, encompassing \textcolor{newtext1}{recruitment campaigns, onboarding, and training and qualification}; (ii) Orchestration and automation, \textcolor{newtext1}{supporting workflow progression, assignment and workload pacing, automated communication, and status and error monitoring}; and (iii) Video management, including \textcolor{newtext1}{video sourcing, privacy and governance support, eligibility screening, and video reprocessing} to yield a curated pool of qualified surgical videos. \textcolor{newtext1}{EAES: European Association for Endoscopic Surgery; IRB: institutional review board.}}
    \label{fig:endoglacier}
\end{figure*}

Managing diverse dataset generation at a global scale requires more than assembling data; it demands infrastructure capable of sustaining continuous, high-quality flows of videos, expert annotations, and coordination across dozens of institutions and contributors. Here, we break down our dataset generation process into three coordinated streams visualized in Figure \ref{fig:endoglacier}: (1) annotator management, (2) video management, and (3) orchestration and automation.

\begin{table*}[ht!]
\centering
\caption{\textcolor{newtext}{Overview of the CVS Challenge pipeline }}
\resizebox{0.8\textwidth}{!}{
\begin{tabular}{p{4cm} p{3.5cm} p{5cm} p{3.5cm}}
\toprule

\textbf{\textcolor{newtext1}{Stage or function}} &
\textcolor{newtext1}{\textbf{Role}} &
\textcolor{newtext}{\textbf{Implementation}} &
\textcolor{newtext}{\textbf{Type}} \\
\midrule

\multicolumn{4}{c}{\textcolor{newtext1}{\textbf{Annotator Management}}} \\
\midrule

\textcolor{newtext1}{Recruitment Campaigns} &
\textcolor{newtext1}{Sourcing expert annotators} &
\textcolor{newtext1}{European Association for Endoscopic Surgery masterclasses, LinkedIn, and X} &
\textcolor{newtext}{Public} \\

\textcolor{newtext1}{Onboarding} &
\textcolor{newtext1}{Introducing the challenge, annotation process, and expected commitment} &
\textcolor{newtext1}{Online onboarding sessions} &
\textcolor{newtext1}{Custom / internal} \\

\textcolor{newtext1}{Training and Qualification} &
\textcolor{newtext1}{Annotator training \newline Resource dissemination \newline Competency evaluation} &
\textcolor{newtext}{MOSaiC} &
\textcolor{newtext}{Proprietary (research)} \\

\midrule
\multicolumn{4}{c}{\textbf{\textcolor{newtext}{Orchestration and Automation (EndoGlacier)}}} \\
\midrule

\textcolor{newtext1}{Workflow progression} &
\textcolor{newtext1}{Progression through ingestion, preprocessing, annotation, and assignment} &
\textcolor{newtext1}{Daily scheduled Python~3 job interfacing with Surgical Safety Technologies and MOSaiC APIs} &
\textcolor{newtext}{Custom / internal} \\

\textcolor{newtext1}{Assignment and workload pacing} &
\textcolor{newtext1}{Controlled allocation of qualified videos among qualified annotators} &
\textcolor{newtext1}{Scheduled Python~3 processes interfacing with MOSaiC} &
\textcolor{newtext}{Custom / internal} \\

\textcolor{newtext1}{Automated communication} &
\textcolor{newtext1}{Notifications, reminders, and requests for manual intervention} &
\textcolor{newtext1}{Automated emails to the organizing team} &
\textcolor{newtext}{Custom / internal} \\

\textcolor{newtext1}{Status and error monitoring} &
\textcolor{newtext1}{Detection of transitions requiring manual action, depleted preannotation queues, and delayed annotation workloads} &
\textcolor{newtext1}{Daily transition polling and separate weekly human-intervention polling} &
\textcolor{newtext}{Custom / internal} \\

\midrule
\multicolumn{4}{c}{\textcolor{newtext1}{\textbf{Video Management}}} \\
\midrule

\textcolor{newtext1}{Video Sourcing} &
\textcolor{newtext1}{Sourcing video contributions} &
\textcolor{newtext1}{European Association for Endoscopic Surgery masterclasses, LinkedIn, and X} &
\textcolor{newtext}{Public} \\

\textcolor{newtext1}{Privacy support} &
\textcolor{newtext}{Surgical video de-identification} &
\textcolor{newtext1}{Surgical Safety Technologies (SST)} &
\textcolor{newtext}{Commercial} \\

\textcolor{newtext1}{Governance support} &
\textcolor{newtext1}{Supporting institutional governance and approval processes} &
\textcolor{newtext1}{Institutional review board templates} &
\textcolor{newtext1}{Custom / internal} \\

\textcolor{newtext1}{Eligibility Screening} &
\textcolor{newtext1}{Eligibility and duplicate-upload checks \newline Preannotation and metadata recording} &
\textcolor{newtext}{MOSaiC} &
\textcolor{newtext}{Proprietary (research)} \\

\textcolor{newtext1}{Video Reprocessing} &
\textcolor{newtext1}{Video clip generation \newline Date--time overlay blurring} &
\textcolor{newtext1}{Python~3 video-processing scripts and Amazon Web Services (AWS) infrastructure} &
\textcolor{newtext1}{Custom / commercial} \\

\bottomrule
\end{tabular}
}
\label{tab:endoglacier}
\end{table*}

\subsection{Annotator Management}

Our fundamental goal was to assemble the largest and most diverse pool of annotators possible. Size was critical to capture the inherent subjectivity in surgical quality assessments robustly, and diversity was essential to minimize bias from individual raters.

To achieve this, we conducted dedicated onboarding sessions that introduced potential annotators to the challenge, its goals, the team, and the commitments expected. Quality control mechanisms were put in place to identify individuals with the expertise, resources, and motivation to meaningfully contribute.

Annotators were provided with expert-consensus protocols, flashcards, and training materials, along with written instructions and video tutorials to familiarize themselves with MOSaiC \citep{mazellier2023mosaic}, a web-based annotation platform developed at IHU-Strasbourg. Beyond providing the annotation interface, MOSaiC enabled blinded scoring of every video, centralized all instructional material for quick reference, and \textcolor{newtext1}{application programming interfaces (APIs)} that allowed automated assignment of videos and annotators, ensuring consistency and efficiency at scale. As a final gate, we implemented a competency-based exam requiring at least 75\% correspondence with expert ratings to ensure conceptual alignment and minimum annotation quality.

\textcolor{newtext}{Through this structured pipeline, 106 potential annotators were contacted or initiated contact, recruited via social media campaigns and in-person onboarding initiatives. Of these, 71 met eligibility criteria and were formally onboarded (2 were excluded due to non-clinical backgrounds). Sixty-seven proceeded to annotation training and the competency examination, of whom 27 passed. Seven candidates did not pass the examination. Of the 27 who passed, 7 subsequently withdrew, resulting in a final qualified pool of 20 expert annotators.}

\textcolor{newtext}{In total, 77 of the original 106 candidates discontinued participation at various stages of the pipeline, most commonly due to time constraints and competing professional commitments.}

\subsection{Video Management}
The primary goal of the video pipeline was to assemble the largest and most diverse dataset of laparoscopic cholecystectomy procedures possible, while ensuring each sample was clinically meaningful and task-relevant for the challenge. The core prediction task in the CVS Challenge is to determine whether the Critical View of Safety (CVS) has been achieved during the 90-second window immediately prior to clipping of the cystic duct or artery, the critical decision point at which misidentification can lead to a bile duct injury.

To support this objective, we launched a broad video sourcing campaign over a three-year period, including social media outreach, in-person recruitment events, and direct institutional engagement. \textcolor{newtext}{Data donors were instructed not to submit data from publicly available cholecystectomy datasets to the challenge.} This effort yielded a total of 1,419 donated surgical videos. We supported contributors with practical tools and resources, including institutional review board (IRB) templates and a custom-built de-identification platform for redacting out-of-body segments that may contain patient or staff identifiers.

All incoming videos were subjected to a manual screening and preannotation phase conducted by members of the organizing team. This process had two main purposes:

\begin{enumerate}
    \item \textbf{Eligibility verification}, based on a predefined set of criteria:
    \begin{itemize}
        \item The video must be of a minimally invasive cholecystectomy (either laparoscopic or robotic).
        \item A continuous 90-second segment prior to clipping of the cystic duct or artery must be available, during which the operative field is clearly visible.
        \item Videos were excluded if they involved a change in surgical strategy due to unsafe conditions, referred to as a \textit{bailout procedure  }, such as conversion to open surgery or subtotal cholecystectomy.
        \item Videos were also excluded if they were incomplete or did not capture the clipping of the cystic structures.
    \end{itemize}
    
    \item \textbf{Preannotation of objective metadata}, with minimal room for interpretation:
    \begin{itemize}
        \item The timestamp of the cystic duct or artery clipping, used to extract the 90-second target segment. \textcolor{newtext}{Note this was defined as the first frame in which the clip is unambiguously visible as applied to the cystic duct or artery.}
        \item The use of adjunctive visualization techniques, such as intraoperative cholangiography (IOC) or indocyanine green (ICG) fluorescence, which may indicate case difficulty or reinforce CVS implementation.
        \item The surgical approach used (laparoscopic vs. robotic), which significantly affects procedural workflow.
    \end{itemize}
\end{enumerate}

To ensure quality and consistency, each video was independently reviewed and preannotated by two raters. In cases of disagreement, a third rater was introduced to adjudicate. This process was repeated iteratively until two consecutive raters produced concordant annotations across all eligibility and metadata criteria.

For privacy compliance, 57 videos required additional reprocessing to blur overlaid date-time stamps. \textcolor{newtext}{Additionally, all videos were manually reviewed to verify the integrity of the 90-second decision window and the effectiveness of de-identification. Videos were excluded if de-identification failed to remove sensitive information or if redaction obscured the operative field within the target segment.} All qualified videos were \textcolor{newtext}{reviewed in their original form,} then clipped to produce 90-second segments corresponding to the critical decision window. This resulted in a final pool of 1,000 high-quality, clinically relevant qualified video clips. \textcolor{newtext}{Importantly, exclusions related to strategy-change scenarios represented a small fraction of the collected dataset (e.g., subtotal $n=6$, converted $n=3$, aborted $n=4$). The majority of exclusions arose from invalid or non-compliant submissions, including videos from unrelated procedures, recordings with overlays or heavy editing, incomplete captures, or low-quality sources such as phone recordings.}

Each qualified clip was subsequently assigned to three independent, qualified annotators, who were asked to assess whether the CVS had been achieved and to rate their confidence in the assessment. This multi-annotator protocol was designed to capture the inherent subjectivity of surgical quality assessment and provide a basis for evaluating inter-rater agreement.

\subsection{EndoGlacier: Orchestration and Automation}

Handling this volume of data, annotators, and institutional contributors over a multi-year period would have been intractable without automation. \textcolor{newtext}{The overall pipeline is composed of }\textcolor{newtext1}{three coordinated flows: annotator management, orchestration and automation, and video management. The annotator and video management flows comprise sequential stages, whereas orchestration and automation comprises cross-cutting functions connecting and monitoring these stages. }\textcolor{newtext}{ These components, along with the orchestration layer coordinating them, are summarized in Table~\ref{tab:endoglacier}.}

\textcolor{newtext}{In practice, EndoGlacier }\textcolor{newtext1}{was implemented as a set of scheduled Python~3 jobs interfacing with the application programming interfaces (APIs) of Surgical Safety Technologies and MOSaiC }\textcolor{newtext}{ to manage data transfer between stages, enforce standardized video preprocessing (e.g., clipping and re-encoding), monitor pipeline status, and communicate with annotators and organizers through automated notifications.} \textcolor{newtext1}{A daily polling job identified and processed videos requiring progression from Surgical Safety Technologies to the MOSaiC preprocessing workspace, from preprocessing to the MOSaiC annotation workspace, and from the annotation workspace to assignment among qualified annotators. When a transition required manual intervention, the system emailed the organizing team with the affected video and the required action. A separate weekly polling job identified whether additional videos needed to be made available to the screening and preannotation team and whether annotators were falling behind their assigned workloads.} \textcolor{newtext}{These mechanisms enabled parallel progression of videos and annotators through the pipeline, while maintaining visibility over system state and allowing timely intervention when required. The orchestration layer also }\textcolor{newtext1}{supported assignment and workload pacing across qualified annotators.}

\textcolor{newtext}{Importantly, the system was designed to be modular }\textcolor{newtext1}{at the workflow level, but its executable implementation was platform-specific.} \textcolor{newtext}{While certain components currently rely on proprietary or commercial solutions, }\textcolor{newtext1}{the reusable elements are the stage sequence, transition logic, and monitoring and intervention points described here. The Python scripts are project-specific adapters to non-public Surgical Safety Technologies and MOSaiC interfaces and are therefore not distributed as a standalone package. Implementing the same workflow with different platforms would require reimplementation of the corresponding platform adapters.}

\section{SAGES CVS challenge}
\label{sec:challenge}
\subsection{Benchmark Task: Assessing the Critical View of Safety}

The Critical View of Safety (CVS) is a universally recommended surgical safety step in laparoscopic cholecystectomy. Its purpose is to ensure that the key aspects of dissection have been performed with sufficient quality to allow conclusive identification of the cystic duct and cystic artery, before proceeding to clip and cut these structures. This step is central to preventing bile duct injuries, a devastating and largely preventable complication.

CVS is composed of three binary criteria, all of which must be met to achieve the critical view:

\begin{itemize}
    \item \textbf{Criterion 1 (C1):} Two and only two tubular structures are visible entering the gallbladder.
    \item \textbf{Criterion 2 (C2):} The hepatocystic triangle is cleared of fat and fibrous tissue.
    \item \textbf{Criterion 3 (C3):} The lower third of the gallbladder is detached from the liver bed.
\end{itemize}

The decision to proceed is most relevant in the final moments prior to clipping and cutting the cystic duct and artery. To reflect this clinical reality, the CVS Challenge focuses assessment on the 90-second window immediately preceding this critical juncture in the procedure.

To balance clinical relevance with annotation effort and modeling tractability, the benchmark is structured such that CVS achievement can be assessed at a granularity of one frame every five seconds within this 90-second window. This design provides both temporal resolution and practical annotation density, aligning the benchmark with the real-world clinical decision context.

\subsection{Dataset Characteristics}

The SAGES CVS Challenge dataset comprises a total of 1,000 laparoscopic cholecystectomy cases, each represented by a 90-second clip taken from the critical decision window preceding clipping of the cystic duct or artery. The dataset is split into a training set of 700 videos and a test set of 300 videos (Table~\ref{tab:dataset_summary}). \textcolor{newtext}{The split was performed randomly, reflecting unselected real-world distributions at this scale.} To enable practical annotation density while preserving temporal resolution, one frame every five seconds (18 frames per video) is annotated throughout the clip.

Annotator confidence, self-reported on a per-clip basis (0--1 scale), shows a mean of 0.64 $\pm$ 0.28 on the training set and 0.58 $\pm$ 0.27 on the test set. As expected, confidence varies across cases, reflecting case complexity and clinical ambiguity.

The dataset captures substantial clinical and technical diversity. 23 countries are represented in the training split and 18 countries in the test split, with an average of 30.43 $\pm$ 46.54 and 16.57 $\pm$ 23.18 videos per country, respectively. Data acquisition was performed across the same 8 device vendors in both splits. A portion of videos lack device metadata (156 in train, 114 in test), but the remaining videos are evenly distributed, with a mean of 68.0 $\pm$ 65.47 videos per device in the training set and 23.25 $\pm$ 24.44 in the test set. \textcolor{newtext}{In addition to labels, the dataset includes selected clinical and technical metadata. Country of origin and device vendor are provided as categorical variables (pseudonymized to preserve privacy). The use of robotic surgery, indocyanine green (ICG), and intraoperative cholangiography (IOC) are provided as binary indicators (yes/no). Missing entries correspond to cases where metadata was not available at the time of submission.}

\begin{figure}[ht]
     \includegraphics[width=1.0\linewidth]{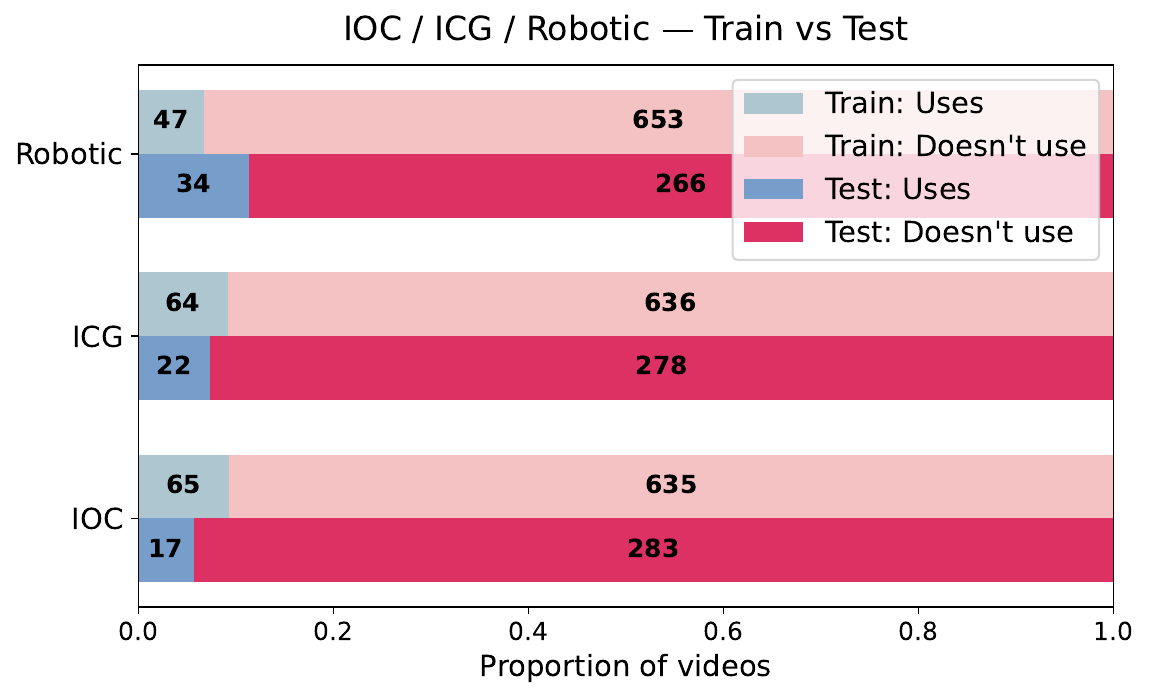} 
    \caption{Distribution of adjunct imaging techniques and surgical platform across training and test sets. Proportion of videos using intraoperative cholangiography (IOC), indocyanine green (ICG) fluorescence imaging, and robotic-assisted surgery. The distribution demonstrates clinical and workflow diversity across the dataset, with variation between training and test splits done randomly to reflect unselected real-world patterns.}
    \label{fig:ioc_icg_robotic_dataset}
\end{figure}

\begin{figure*}[t]
    \centering
    \includegraphics[width=0.9\textwidth]{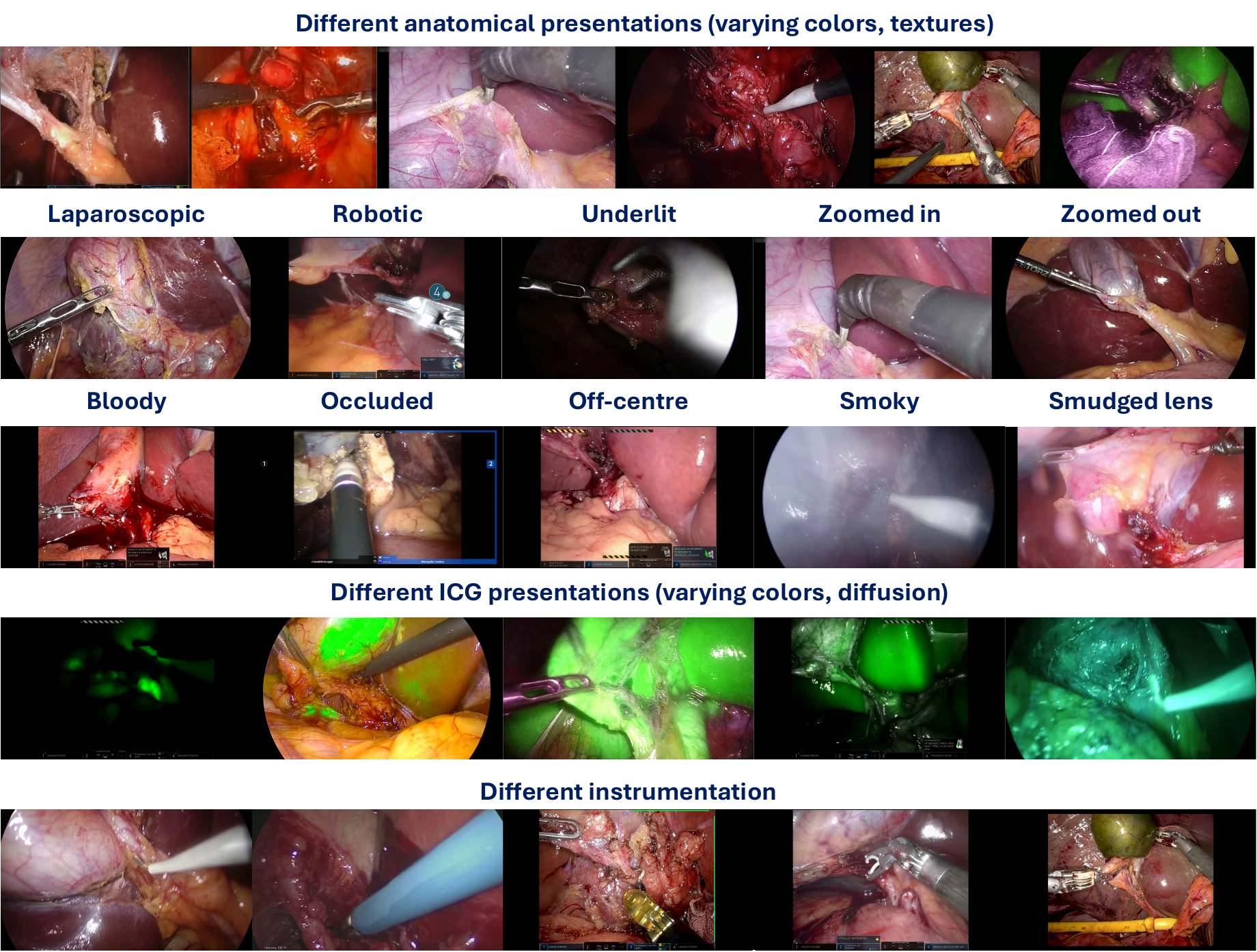}
    \caption{\textcolor{newtext1}{A snapshot of visual and procedural diversity in the SAGES CVS Challenge dataset. The examples illustrate variation in anatomical and tissue appearance, surgical platform, instrument type, illumination, field of view, framing, and indocyanine green (ICG) fluorescence appearance, as well as conditions impairing visibility, including blood, occlusion of the anatomy by a surgical instrument, smoke, and lens contamination.}}
    \label{fig:dataset_diversity}
\end{figure*}

The use of robotic platforms, Indocyanine Green (ICG) fluorescence, and Intraoperative Cholangiography (IOC) varies across the dataset and is summarized in Figure~\ref{fig:ioc_icg_robotic_dataset}. Most procedures are performed laparoscopically (653/700 in train, 266/300 in test), with ICG and IOC used in a minority of cases, providing an opportunity to examine model robustness across clinically heterogeneous workflows. \textcolor{newtext1}{Beyond these recorded metadata, the dataset exhibits substantial visual and procedural variability relevant to automated CVS assessment. Figure~\ref{fig:dataset_diversity} illustrates differences in anatomical and tissue appearance, surgical platform and instrumentation, illumination, field of view, and framing, as well as conditions impairing visibility, including blood, surgical instruments occluding the anatomy, smoke, and lens contamination. It also demonstrates variation in the appearance of indocyanine green (ICG) fluorescence across cases.}

Annotation of Critical View of Safety (CVS) achievement is performed for each frame along three binary criteria (C1, C2, C3). The dataset captures not only frame-level classification but also annotator agreement, which is known to vary in subjective quality assessment tasks. We distinguish between full agreement (all 3 annotators concur) and partial agreement (2 out of 3 annotators agree). Such variation is expected and informative: all annotators were screened through structured training and competency validation to minimize sources of bias unrelated to inherent subjectivity, including lack of alignment with the protocol, inattention, or annotation fatigue.

\begin{figure*}[ht]
    \includegraphics[width=1.0\linewidth]{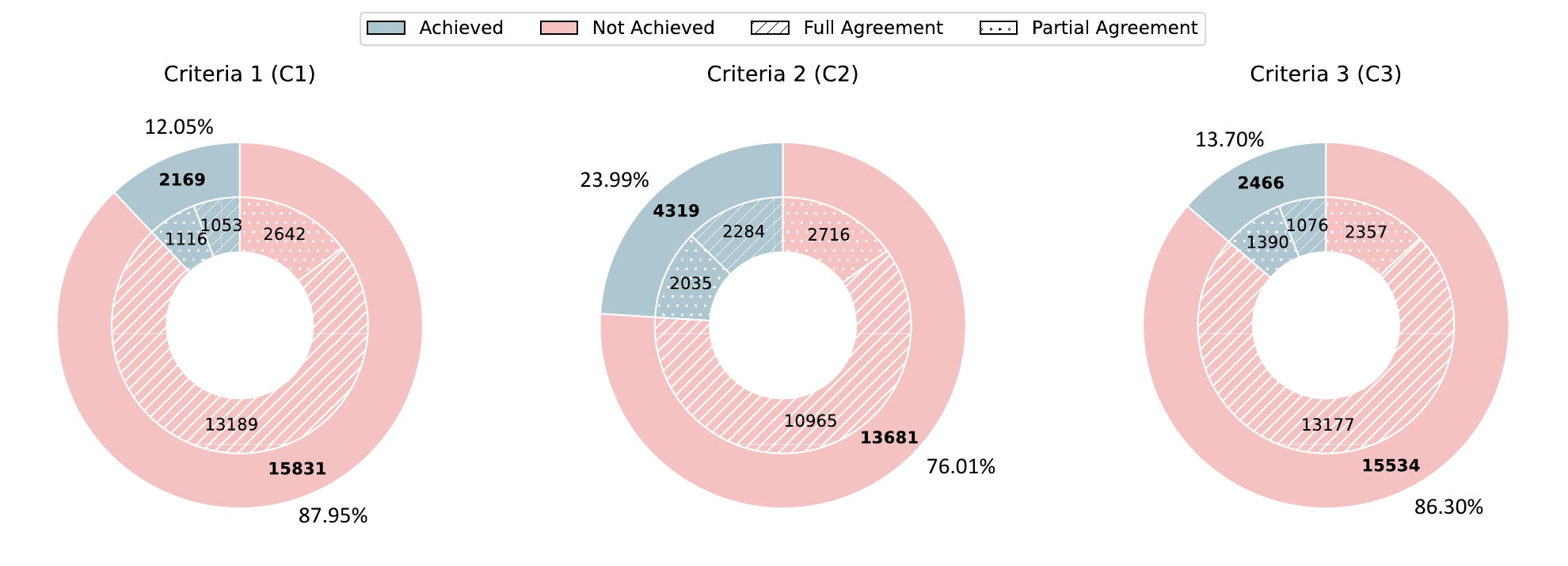} 
    \caption{Distribution of CVS assessment outcomes and annotator agreement. For each CVS criterion (C1, C2, C3), we report the number of clips rated as ``Achieved" vs. ``Not Achieved," and the corresponding level of annotator agreement (full agreement across all three annotators vs. partial agreement). This highlights both the subjective nature of CVS assessment and the distribution of clinical outcomes in the dataset.}
    \label{fig:cx_dataset}
\end{figure*}

Figure~\ref{fig:cx_dataset} presents the distribution of frame-level achievement rates and agreement levels for each criterion. As expected, C2 (clearing the hepatocystic triangle) shows the highest frame-level achievement rate, while C1 (two and only two tubular structures visible) and C3 (lower third of gallbladder detached) are achieved less frequently. \textcolor{newtext}{Across all criteria, ``Not Achieved'' states dominate and are associated with higher levels of  agreement, suggesting that the absence of CVS criteria is more consistently identified than their presence. In contrast, ``Achieved'' states exhibit a higher proportion of partial agreement, particularly for C1 and C3, reflecting increased ambiguity in confirming full criterion satisfaction. Overall, these patterns highlight the asymmetric and criterion-dependent nature of inter-rater variability in CVS assessment, reinforcing the importance of modeling uncertainty in this task.}

At the video level, annotators assessed whether each CVS criterion was achieved during the 90-second window. Across the dataset:
\begin{itemize}
    \item \textbf{C1}: 413 videos achieved, 587 not achieved
    \item \textbf{C2}: 600 videos achieved, 400 not achieved
    \item \textbf{C3}: 395 videos achieved, 605 not achieved
\end{itemize}

Overall, the dataset reflects the real-world variability, subjectivity, and technical heterogeneity that models deployed in surgical environments will encounter. By capturing this complexity, the CVS Challenge provides a benchmark that goes beyond technical optimization and instead tests the robustness and clinical relevance of AI-based surgical quality assessment.

\begin{table}[ht]
\centering
\caption{Summary statistics of the SAGES CVS Challenge dataset. Key characteristics of the video dataset and annotations used for training and evaluation. Annotator confidence is self-reported by annotators on a per-clip basis. Country and device metadata are reported separately for training and test splits to highlight dataset diversity; \textcolor{newtext}{missing entries correspond to cases where this information was not reported.}}
\label{tab:dataset_summary}
\resizebox{\columnwidth}{!}{%
\begin{tabular}{llllll}
\toprule
\multicolumn{4}{l}{\textbf{Number of Videos (90-second clips)}} & \textbf{Total} & \textbf{1000} \\
\cmidrule(lr){5-6}
\multicolumn{4}{l}{}                                           & Train          & 700           \\
\multicolumn{4}{l}{}                                           & Test           & 300           \\
\midrule
\multicolumn{4}{l}{\textbf{Number of frames annotated per video}} & \multicolumn{2}{l}{\textbf{18}} \\
\multicolumn{4}{l}{Annotation frequency} & \multicolumn{2}{l}{1 frame every 5 seconds} \\

\midrule
\multicolumn{6}{l}{\textbf{Annotator Confidence (per clip, 0--1 scale)}} \\
\midrule
\multirow{2}{*}{} & \textbf{Split} & \textbf{\textcolor{newtext1}{Minimum}} & \textbf{\textcolor{newtext1}{Maximum}} & \textbf{Mean} & \textbf{\textcolor{newtext1}{Standard Deviation}} \\
                  & Train          & 0            & 1            & 0.64          & 0.28        \\
                  & Test           & 0            & 1            & 0.58          & 0.27        \\
\midrule
\multicolumn{4}{l}{\textcolor{newtext}{Mean \# annotated cases per annotator (± \textcolor{newtext1}{Standard Deviation)}}} & \multicolumn{2}{l}{\textcolor{newtext}{150.0 ± 99.4}} \\
\midrule
\multicolumn{2}{l}{\textbf{Country metadata}} & \multicolumn{2}{c}{\textbf{Train}} & \multicolumn{2}{c}{\textbf{Test}} \\
\cmidrule(lr){3-4} \cmidrule(lr){5-6}
\multicolumn{2}{l}{Total number of countries}      & \multicolumn{2}{c}{23} & \multicolumn{2}{c}{18} \\
\multicolumn{2}{l}{Videos from unknown country}    & \multicolumn{2}{c}{0}  & \multicolumn{2}{c}{0} \\
\multicolumn{2}{l}{Mean \# videos per country (± SD)} & \multicolumn{2}{c}{30.43 ± 46.54} & \multicolumn{2}{c}{16.57 ± 23.18} \\
\midrule
\multicolumn{2}{l}{\textbf{Device metadata}} & \multicolumn{2}{c}{\textbf{Train}} & \multicolumn{2}{c}{\textbf{Test}} \\
\cmidrule(lr){3-4} \cmidrule(lr){5-6}
\multicolumn{2}{l}{Total number of device vendors}    & \multicolumn{2}{c}{8} & \multicolumn{2}{c}{8} \\
\multicolumn{2}{l}{Videos from unknown device}        & \multicolumn{2}{c}{156} & \multicolumn{2}{c}{114} \\
\multicolumn{2}{l}{Mean \# videos per device (± SD)}     & \multicolumn{2}{c}{68.00 ± 65.47} & \multicolumn{2}{c}{23.25 ± 24.44} \\
\bottomrule
\end{tabular}
} 
\end{table}

\subsection{Challenge Tasks and Evaluation Metrics}

Participants were evaluated on a held-out test set of 300 laparoscopic cholecystectomy videos submitted through the Grand Challenge platform using Docker-based containers. Test videos and annotations were withheld, and all models were required to operate causally using only past and current frames at inference time. Each submission was evaluated across three subchallenges designed to probe different facets of the modeling problem. General updates regarding the challenge were made available through the challenge website \footnote{\url{https://www.cvschallenge.org/}}. Members of the organizing team and their immediate research groups were not permitted to participate in the Challenge.

\paragraph{Subchallenge A: CVS Achievement}
This task evaluated classification performance in predicting the most common (majority) expert assessment for each of the three CVS criteria. The evaluation metric was mean average precision (mAP), computed per frame and averaged across criteria. The ground truth for each frame was defined as:
\begin{equation}
y = \operatorname{mode}(\{ l_i \}), \quad i \in \{1, 2, 3\}
\end{equation}

where \( l_i \) is the binary label provided by annotator \( i \)\textcolor{newtext}{, and \(\operatorname{mode}(\cdot)\) denotes the most frequently occurring value in the set.}

\paragraph{Subchallenge B: Uncertainty Quantification}
This task tested whether models could express uncertainty in line with annotator disagreement and confidence. Participants submitted a probability between 0 and 1 for each frame and criterion. The evaluation metric was the Brier Score (BS), calculated using a soft label incorporating annotator confidence. \textcolor{newtext}{This soft label was generated by aggregating annotator labels weighted by their self-reported confidence. Specifically, each annotator’s binary label is modulated by a confidence score, with low-confidence assessments contributing values closer to 0.5 and high-confidence assessments closer to 0 or 1. These values are then averaged across annotators to produce a continuous calibration target that reflects both consensus among raters and the confidence of each rater.}
\begin{equation}
\tilde{y} = \frac{1}{3} \sum_{i=1}^{3} \left( 0.5 + (l_i - 0.5)  c_i \right)
\end{equation}
where \( l_i \) is the binary label and \( c_i \in [0, 1] \) is the confidence score provided by annotator \( i \). The Brier Score was computed per frame, then averaged across frames and criteria.

\paragraph{Subchallenge C: Domain Shift Robustness}
This task evaluated model generalization under real-world distribution shifts~\citep{ben2010theory,shao2024supervised}. Several variant test sets were constructed from the full test split to introduce differences in imaging modality (IOC, ICG), surgical platform (robotic vs. laparoscopic), device type, country of origin, and annotator confidence. mAP was calculated per variant set, \textcolor{newtext}{denoted \(\text{mAP}_v\),} and the final \textcolor{newtext1}{Domain Robustness Score (DRS)} was defined as:
\begin{equation}
\text{Domain Robustness Score} = \min_{\text{variant } v \in \mathcal{V}_{\text{10--100}}} \text{mAP}_v
\end{equation}
\textcolor{newtext}{where \(\mathcal{V}_{\text{10--100}}\) denotes the set of variant test splits after excluding the bottom 10th percentile of mAP values across all variants to reduce sensitivity to extreme outliers. Here, each \(\text{mAP}_v\) corresponds to the performance on a specific variant test set \(v\).} Ground truth labels were again computed using majority vote:
\begin{equation}
y = \operatorname{mode}(\{ l_i \}), \quad i \in \{1, 2, 3\}
\end{equation}

\subsection{Challenge Design and Submission Framework}
\begin{figure*}[ht]
         \centering
\includegraphics[width=0.9\linewidth]{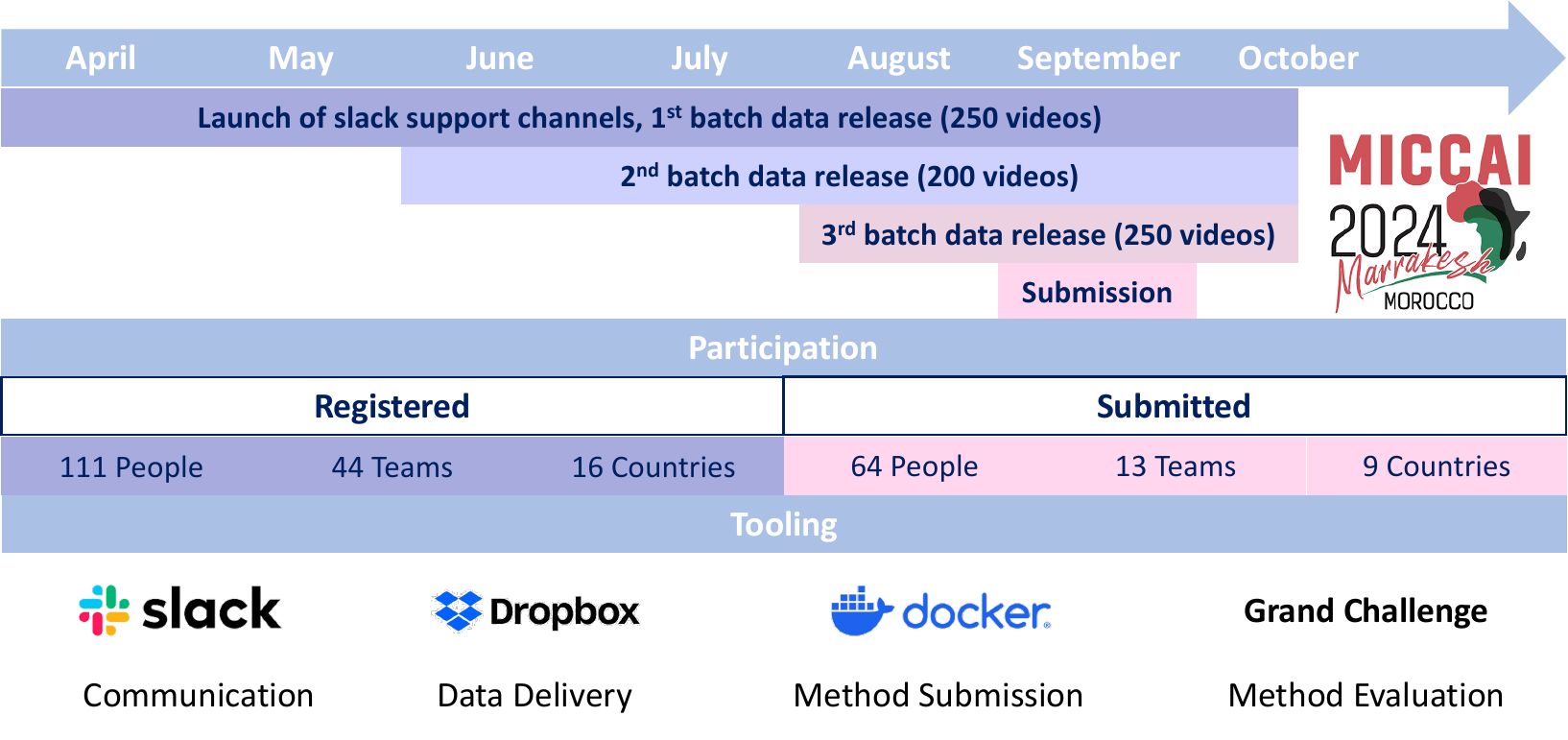} 
    \caption{Overview of the SAGES CVS Challenge. The top panel shows dataset release and timeline, the middle panel summarizes team participation, and the bottom panel highlights the core tooling stack used to manage communication, data delivery, and evaluation.}
    \label{fig:challenge_design}
\end{figure*}

The SAGES CVS Challenge was designed to balance clinical realism with reproducibility and ease of participation (Fig. \ref{fig:challenge_design}). Submissions were made through the Grand Challenge platform\footnote{\url{https://cvs-challenge.grand-challenge.org/}}, using Docker containers to ensure standardized evaluation. Participants were required to implement causal models, making predictions at each timepoint using only the current and preceding frames, in line with intraoperative constraints. No restrictions were placed on external data sources, as long as they were publicly accessible, ensuring a fair playing field across teams. \textcolor{newtext}{Participation was solicited through a combination of public and targeted outreach. This included announcements via professional networks (e.g., SAGES and MICCAI communication channels), social media campaigns (LinkedIn, X), and direct outreach to research groups active in surgical data science and medical video analysis.}

To support participation, an open-source submission template was provided, including example code, a test video, and scripts for local building, testing, and packaging. Each model was expected to ingest a single 90-second, 1 \textcolor{newtext1}{frame per second (fps)} video and output predictions for all three CVS criteria at each of the 90 frames. 

Evaluation was automated via Grand Challenge’s infrastructure. Submissions were uploaded as Docker containers, registered and validated on the platform, and then tested against the held-out dataset. Participants were required to test locally using the provided inference interface, and could optionally validate on-platform using an example input video. \textcolor{newtext}{To support troubleshooting, participants also had access to dedicated communication channels, including a monitored Slack workspace and support email, allowing environment- and submission-related issues to be resolved during the validation period.} Participants submitted a single Docker container, which was evaluated across all three subchallenges. This design reflected real-world deployment constraints: in surgical AI systems, raw predictive performance, reliability under distribution shifts, and appropriate uncertainty estimation are all critical, yet not inherently aligned. Optimizing for one may compromise another. Requiring a single model to be evaluated across all criteria forced participants to navigate these trade-offs, as would be required for clinically deployable systems.

\subsection{Awards}
Monetary prizes were awarded to the top 3 ranking teams for each of the 3 subchallenges and announced at MICCAI 2024. The overall winner across the 3 subchallenges was additionally awarded an NVIDIA IGX edge AI platform.
\section{Methods}

\subsection{Baseline: LG-DG}
As a baseline for our challenge, we adapted LG-DG \citep{satyanaik2024optimizing}, a recently proposed object-centric classification model optimized for domain generalization in surgical video. LG-DG builds on a previous latent graph framework (LG-CVS, \citep{murali2023latent}) by explicitly disentangling semantic, visual, and image-level features through a combination of architectural and loss-based design choices.

The model operates in two stages: first, a Mask-RCNN \citep{he2017mask} detector identifies and localizes surgical tools and anatomical structures, encoding their spatial, semantic, and visual properties as nodes and edges in a latent graph. This graph is passed to a graph neural network-based classification head for downstream prediction of CVS criteria. To improve robustness, the model is trained with an auxiliary reconstruction objective and a disentanglement loss that regularizes the predictions from masked graph variants, emphasizing different feature categories.

For this challenge, we retrained LG-DG using spatial supervision from the Endoscapes2023 dataset and CVS annotations from the SAGES CVS Challenge training set.

\subsection{\textcolor{newtext}{Baseline: ConceptNet}}

\textcolor{newtext}{To provide an alternative modeling perspective, we include a second baseline model based on a concept graph neural network architecture. This approach adapts the Concept Graph Neural Network (ConceptNet) framework~\citep{ban2023concept} to the CVS assessment task.}

\textcolor{newtext}{Each input frame is processed using a DINO-pretrained Vision Transformer (ViT-B/16) backbone~\citep{caron2021emerging} to extract visual features. These features are then integrated within a temporal concept graph, composed of three nodes corresponding to the CVS criteria (C1–C3), and a hyperedge representing overall CVS achievement, defined as the conjunction of the three criteria at the frame level.}

\textcolor{newtext}{Each concept node maintains a latent state through a recurrent unit (LSTM), enabling temporal aggregation of evidence. At each time step, the hidden states of the concept nodes are combined through a learned hyperedge encoder, producing a relational representation of the overall CVS state. This representation is updated via a relation-level recurrent unit, which incorporates both visual features and a learned global context vector. The updated relation state is then propagated back to the individual concept nodes, which subsequently update their own states.}

\textcolor{newtext}{Prediction heads are attached to each concept node and to the hyperedge, producing probabilities for each criterion (C1–C3) as well as for overall CVS achievement. Temporal modeling is performed over a fixed history of eight frames.}

\textcolor{newtext}{In contrast to the original ConceptNet formulation, which models fine-grained anatomical entities, this adaptation directly uses the CVS criteria as the conceptual nodes, aligning the architecture with the evaluation protocol of the challenge.}

\begin{sidewaystable*}[p]
  \centering
  \scriptsize
  \caption{Summary of teams, architectures and training details.}
  \label{tab:challenge-summary}
  \resizebox{0.9\textwidth}{!}{
  \begin{tabularx}{\textwidth}{ 
      >{\raggedright\arraybackslash}c  
      >{\raggedright\arraybackslash}p{1.5cm}  
      >{\raggedright\arraybackslash}p{4cm}  
      >{\raggedright\arraybackslash}p{7.5cm}  
      >{\raggedright\arraybackslash}p{2cm}  
      >{\raggedright\arraybackslash}X  
      >{\raggedright\arraybackslash}p{5cm}  
    }
    \toprule
    \# & Team           & Affiliation(s)                                        & Architecture                                                                                                                        & Temporal Component   & Auxiliary Objective                                                                                                      & Pretraining                                                                                      \\
    \midrule
    1  & TUE-VCA        & Eindhoven University of Technology (Netherlands)      & PVT-v2 (x5 ensemble)                                                                                                                 & None                 & L2 reconstruction loss                                                               & Self-supervised (DINO on SurgeNet and GenSurgery datasets)                                 \\
    2  & Farm           & Stanford University (USA)                             & DINOv2 (giant, large), SigLIP, ConvNeXt2-L, InternImage                                                                             & Stacked dilated TCN  & None                                                                                                                & Self-supervised (DINO, SigLIP)                                                                   \\
    3  & CVS HUST       & Huazhong University of Science and Technology (China) & ConvNeXt + MLP                                                                                                                       & LSTM                 & \(\ell_1\) loss for confidence prediction                                                                             & Supervised (natural image weights)                                                               \\
    4  & SDS-HD         & German Cancer Research Center (DKFZ) (Germany)        & EVA02-Large Transformer                                                                                                              & None                 & Segmentation loss on Endoscapes-pretrained YOLOv8 pseudolabels                                                       & Self-supervised (MoCov2 on Cholec80, HeiChole, Endoscapes)                                         \\
    5  & SRV-WEISS      & University College London (UK)                        & EndoViT + Dense Prediction Transformer, EfficientNet, ConvNeXt                                                     & None                 & Segmentation + classification fusion                                                                                                                 & Self-Supervised (EndoViT) + Supervised (EndoViT on Endoscapes segmentation; EfficientNet on Endoscapes classes)              \\
    6  & Caresyntax      & Caresyntax                                                     &  DenseNet-121 (5× ensemble)  & None                                                                                                         & None                 & Supervised (ImageNet weights) \\
    7  & FightTumor     & University Health Network (Canada)                    & ConvNeXt-B                                                                                                                           & None                 & None                                                                                                                 & Supervised (ImageNet weights)                                                                    \\
    8  & HUFT-MedIA     & Hefei University of Technology (China)                & Vision Transformer (ViT) + MLP                                                                                                       & None                 & None                                                                                                                 & Supervised (ImageNet weights)                                                                    \\
    9  & IRCV-URV       & Universitat Rovira i Virgili (Spain)                  & EfficientNet-B5 + FPN + spatial attention (5x ensemble)                                                                                           & None                 & None                                                                                                                 & Supervised (ImageNet weights)                                                                    \\
    10 & mmll           & NAAMII (Nepal)                                        & \begin{tabular}[t]{@{}l@{}}1. ResNet-50 + Mask R-CNN + Transformer encoder (temporal 10-frame)\\ 2. ResNet-50 + Mask R-CNN + Transformer encoder (single frame tokens)\\ 3. LG-CVS\end{tabular} & Transformer                 & Reconstruction Loss, MSE to average rater score                                                                                                                 & Supervised (Endoscapes-BBox201)                                          \\
    11 & Pandas         & ChengDu Withai Innovations Technology Company (China)                                     & ConvNeXt V2                                                                                                                          & None                 & SimCLR-style contrastive loss (NT-Xent)                                                                              & Supervised (ImageNet weights)                                                                    \\
    12 & Theator        & Theator (Israel)                                         & EVA02-Large Transformer (6× model soup ensemble)                                                                                     & Transformer          & Log-cosh for confidence estimation, BCE for video-level CVS labels, Focal loss for pre‐temporal component classification        & Supervised (ImageNet-1k weights)                                                                 \\
    13 & Transformers   & University Hospital Cologne (Germany)                 & U-Net (EfficientNet-B0) + Vision Transformer (ViT)                                                                                     & None                 & Segmentation                                                                                                                 & Supervised (Endoscapes for U-Net)                                              \\
    \bottomrule
  \end{tabularx}
  }
  \label{tab:method_description}
\end{sidewaystable*}

\subsection{Competing methods}
In addition to the baseline model, LG-DG, 13 teams submitted competing methods to the SAGES CVS Challenge. The goal was to predict per-frame achievement of the three CVS criteria using only the labeled 700-video training set. All teams developed their own training pipelines and architectures and submitted Dockerized inference code, which was automatically evaluated across all subchallenges on a hidden 300-video test set. 

The diversity of approaches submitted reflects the multi-faceted nature of the task and the real-world complexity of surgical video understanding. Teams variously framed the problem as framewise classification, sequence modeling, or multitask learning; built on top of pretrained image encoders or segmentation models; and handled label uncertainty through loss design, label smoothing, or ensemble calibration. Several groups also integrated auxiliary tasks or domain adaptation strategies to improve generalization. Many of the top-performing methods combined multiple strategies through architectural design and ensembling.

Below, we describe the methodological design of each submission, with attention to key architectural components, auxiliary objectives, supervision strategies, and other distinctive modeling choices. For an overview of each method’s high-level design choices, see Table~\ref{tab:challenge-summary}.

\team{Team TUE-VCA}
{Scalable CVS Classification via Self-Supervised Pretraining and Semi-Supervised Distillation}
{\textcolor{newtext}{This method uses a PVT-v2 backbone \citep{wang2022pvt} for frame-level CVS classification, combining self-supervised pretraining, semi-supervised distillation, and ensembling without temporal modeling.} The TUE-VCA team addressed the challenge of class imbalance and label uncertainty through a two-stage training approach combining large-scale self-supervised pretraining and semi-supervised knowledge distillation. A Pyramid Vision Transformer v2 (PVT-v2)  backbone was first pretrained using DINO on SurgeNet, a collection of over 4 million unlabeled frames from public surgical datasets \citep{jaspers2024exploring}, and the GenSurgery dataset, which compiles 680 hours of YouTube videos depicting general surgery procedures \citep{schmidgall2024general}. This self-supervised model was then fine-tuned on the labeled training set from the SAGES CVS Challenge.

\begin{figure}[ht]
    \includegraphics[width=1.0\linewidth]{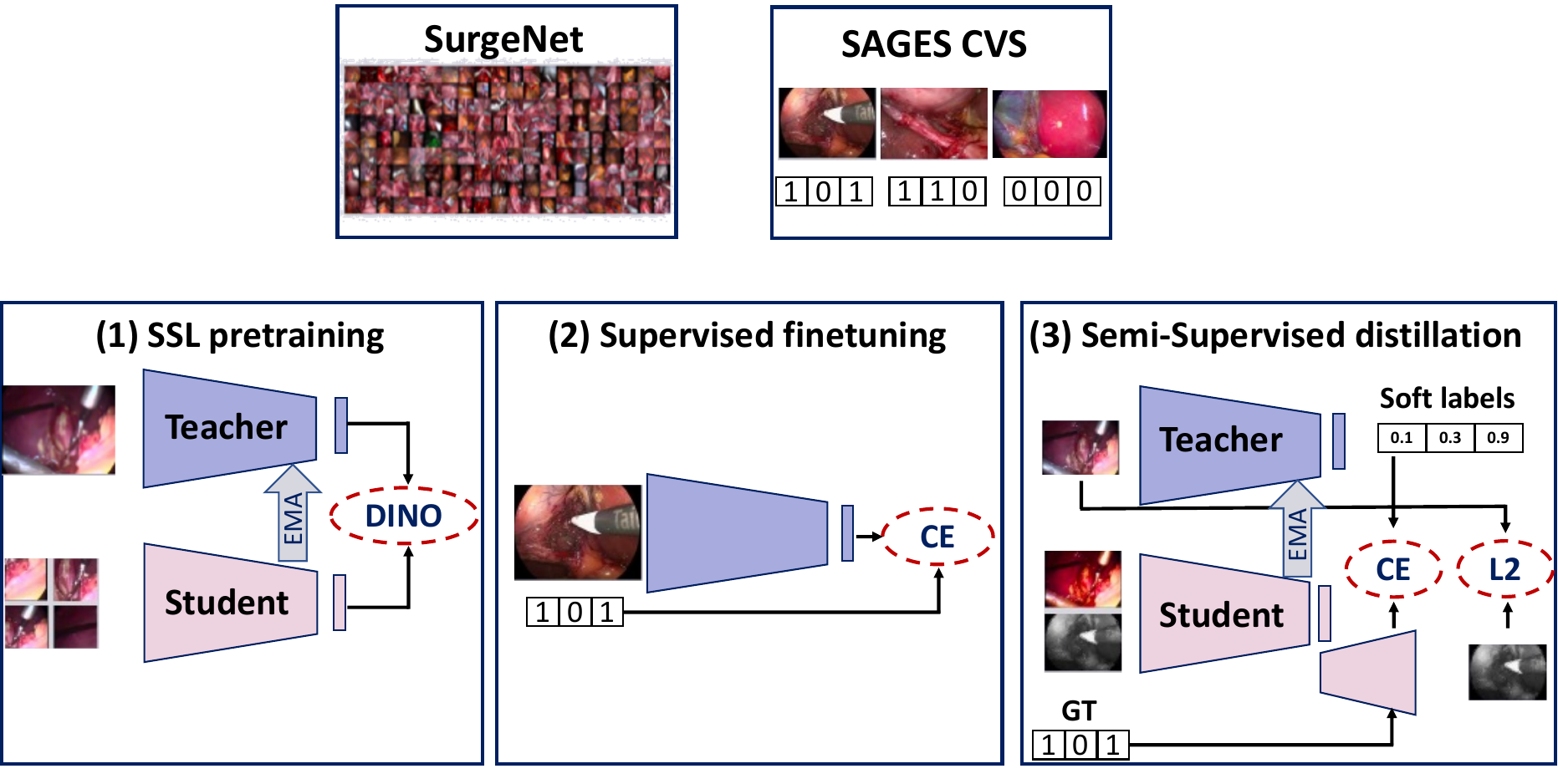} 
    \caption{Overview of various stages of the TUE-VCA method.}
    \label{fig:tue_vca}
\end{figure}

To further improve sensitivity to rare positive cases, the team employed a student-teacher distillation framework. A teacher model trained solely on labeled data was used to identify confident positive frames within the unlabeled data pool. These pseudo-labeled frames were then incorporated into training a student model on both labeled and pseudo-labeled data. Strong augmentations and a diminishing auxiliary reconstruction loss were applied to improve generalization and representation robustness.

The final model operated at the frame level with no temporal modeling. An ensemble of five PVT-v2 models was used for inference, and temperature scaling was applied to calibrate the sigmoid output probabilities following \cite{kumar2022calibrated}. To better account for skewed label distributions, the model was trained using label sampling.

}

\team{Team Farm}
{Efficient Adaptation of Vision Foundation Models with TCNs}
{\textcolor{newtext}{This method uses multiple vision foundation model backbones with parameter-efficient fine-tuning for frame-level CVS classification, with temporal modeling applied in a subset of ensemble members.} Multiple pretrained image backbones—including DINOv2 (giant and large) \citep{oquab2023dinov2}, SigLIP \citep{zhai2023sigmoid}, InternImage \citep{wang2023internimage}, and ConvNeXt2 \citep{woo2023convnext}—were adapted using Low-Rank Adaptation (LoRA) \citep{hu2022lora}, allowing only a small subset of weights to be fine-tuned on the CVS Challenge training set. Some of these backbones were paired with a Temporal Convolutional Network (TCN) \citep{bai2018empirical} to aggregate frame-level features into temporally-aware representations. Models were trained using both full-sequence (90-frame) and shorter-sequence (18-frame) settings, the latter approximating 1 fps inference using overlapping sliding windows.

\begin{figure}[ht]
    \includegraphics[width=1.0\linewidth]{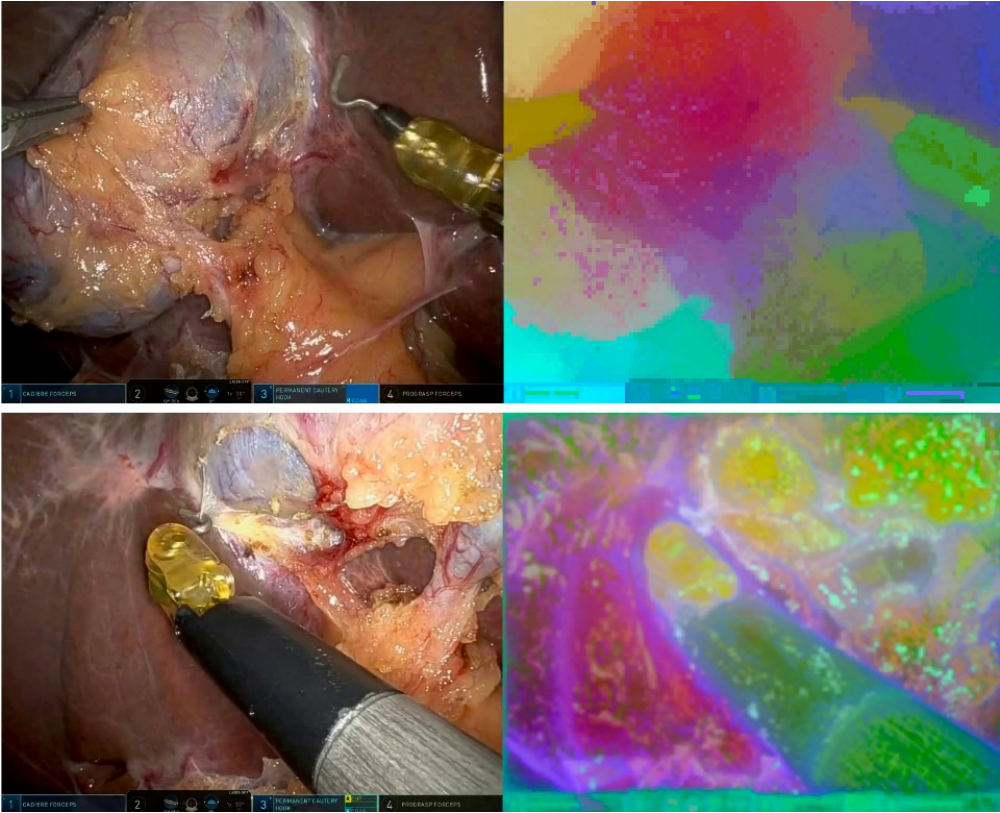} 
    \caption{Overview of various stages of the FARM method. Foundation models, despite being trained primarily on natural images, extract informative features from surgical video frames. By virtue of their large size and extensive pretraining datasets, they produce robust representations of distinct objects and structures - instruments, gallbladder, liver, connective tissue, and other recurring patterns. These representations can be visualized using UMAP \citep{mcinnes2018umap} (top) and PCA (bottom).}
    \label{fig:FARM}
\end{figure}

The final ensemble includes seven models: (1) DINOv2-giant with 90-frame TCN; (2) frame-level fine-tuned DINOv2-giant; (3) SigLIP with 18-frame TCN; (4) DINOv2-large with 18-frame TCN; (5) DINOv2-giant with 18-frame TCN; (6) ConvNeXt2-L with 18-frame TCN; and (7) frame-level fine-tuned InternImage. All models output sigmoid probabilities for each CVS criterion at each annotated frame, trained with binary cross-entropy loss.
}

\team{Team CVS HUST}
{Cyclic CNN-LSTM Training with Bidirectional Parameter Sharing}
{\textcolor{newtext}{This method uses a ConvNeXt-based two-stream architecture combining frame-level and video-level supervision through LSTM-based temporal modeling and an auxiliary confidence objective.} The approach alternates between two training modes: (1) a frame-level CNN stream trained on annotated frames, and (2) a video-level CNN-LSTM stream trained on full 90-frame clips. After each batch, CNN parameters are transferred between the two streams in both directions, promoting mutual refinement across granularities.

ConvNeXt \citep{liu2022convnet}, pretrained on natural images, serves as the backbone. Its first three stages are frozen during training, while later stages are fine-tuned. The temporal stream adds a single-layer LSTM to the CNN backbone. Binary cross-entropy loss is used for classification, and L1 loss is used to supervise confidence scores. The model predicts frame-level outputs for all three CVS criteria. 

\begin{figure}[ht]
    \includegraphics[width=1.0\linewidth]{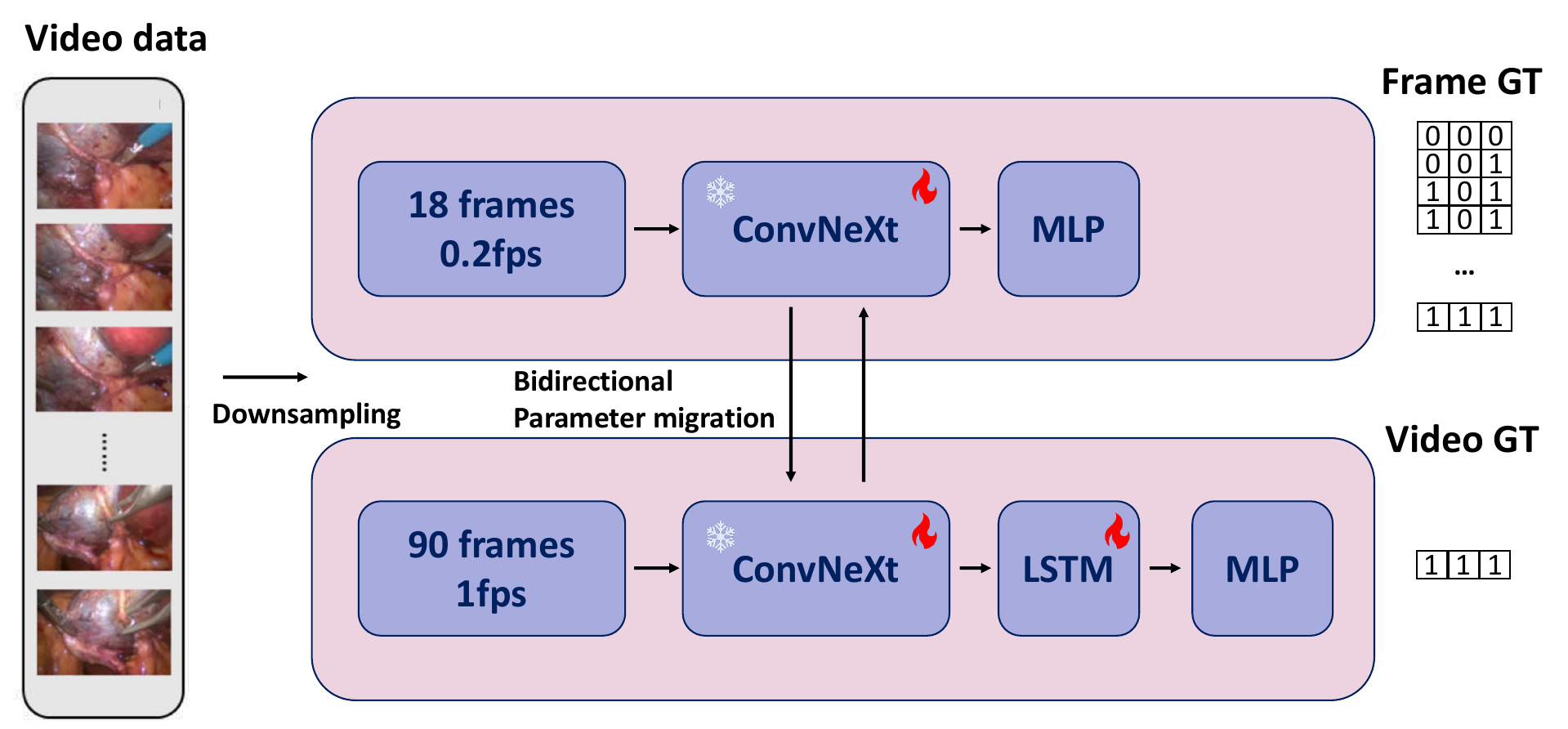} 
    \caption{Overview of various stages of CVS HUST method.}
    \label{fig:cvs_hust}
\end{figure}
}

\team{Team SDS-HD}
{Multitask Transformer with Pseudo-Labels and Uncertainty-Aware Training}
{This method formulates CVS prediction as a multitask learning problem using the EVA02-Large Vision Transformer \citep{fang2024eva} as backbone. The model simultaneously learns to classify CVS criteria and perform segmentation, with the two objectives optimized jointly using a weighted combination of binary cross-entropy, Dice, and Lovász losses. Classification is supervised using a smoothed version of the mean rater score per frame. Specifically, rater labels of {0, 0.33, 0.66, 1} are mapped to {0.15, 0.3975, 0.645, 0.90} using a linear label smoothing function to mitigate overconfidence and reflect annotator disagreement.

Segmentation is guided by pseudo-labels from a YOLOv8 model pretrained on Endoscapes \citep{yolov8_ultralytics, murali2023endoscapes}, which are used during training but not inference. In parallel, a second EVA02-Large model pretrained with MoCoV2 \citep{chen2020improved} on Cholec80 \citep{twinanda2016endonet}, HeiChole, and Endoscapes datasets was trained with the same smoothed labels, forming a complementary representation stream. Finally, both models were used to generate 1 fps pseudo-labels for the training set.

The final submission ensembles five models trained via 5-fold cross-validation on the 700 training videos (140 per fold), grouped by video and stratified by CVS prevalence.
}

\team{Team SRV-WEISS}
{Hybrid Segmentation–Classification Pipeline for CVS Prediction}
{This method combines segmentation- and classification-based representations to predict CVS criteria. The architecture uses a Dense Prediction Transformer (DPT)\citep{Ranftl_2021_ICCV} with an SSL-pretrained EndoViT encoder \citep{batic2024endovit} for anatomical and tool segmentation, and an EfficientNet classifier for high-level recognition of surgical elements. Both models were additionally pretrained on subsets of the Endoscapes dataset and fine-tuned on the challenge task. For each input frame, segmentation masks from DPT are concatenated with the RGB image and passed through a ConvNeXt-tiny encoder, while EfficientNet outputs six class probabilities corresponding to anatomical and tool categories. These outputs are fused and processed by a two-layer multilayer perceptron (MLP) to generate final sigmoid predictions for each CVS criterion.

To adapt to the expanded input dimensionality caused by concatenated segmentations, the ConvNeXt input layer was modified using a weight averaging strategy \citep{wang2016temporal}. All predictions were made per-frame, and training was fully supervised using a weighted binary cross-entropy loss. The weighting accounted for both the prevalence of each class and the annotator's confidence, i.e., an annotator-confidence-aware label distribution. The DPT weights were frozen during training to reduce overfitting and stabilize learning.
}

\team{Team Caresyntax}
{DenseNet Ensemble for Multi-Criteria CVS Prediction}
{This method frames CVS assessment as a multi-label classification task and employs an ensemble of five independently trained DenseNet-121 \citep{huang2017densely} models to improve robustness and generalization. Each model is fine-tuned from ImageNet weights and adapted for multi-label prediction by modifying the final classification head to output probabilities for the three CVS criteria.

To address class imbalance, the team used weighted binary cross-entropy loss. During inference, predictions from the five models were averaged to form the final output. All predictions were made independently for each frame, with no additional temporal modeling, attention mechanisms, or auxiliary tasks. Data augmentation included affine transformations, horizontal flipping, and color jittering. This setup prioritized architectural simplicity and training stability while leveraging ensemble diversity for better performance across criteria.
}

\team{Team FightTumor}
{ConvNeXt-Baseline with AutoAugment for CVS Prediction}
{This method adopts a straightforward multi-label classification architecture consisting of a ConvNeXt-B backbone and a simple multilayer perceptron (MLP) head. The model outputs sigmoid probabilities for each of the three CVS criteria per frame. Training is fully supervised using a weighted binary cross-entropy loss to address class imbalance.

To enhance generalization, the team applied AutoAugment \citep{cubuk2019autoaugment} as the primary data augmentation strategy. The model was initialized with ImageNet-pretrained weights, and no auxiliary losses, attention mechanisms, or temporal modeling were used. All predictions are generated independently for each frame using a single-pass, end-to-end model.
}

\team{Team HFUT-Media}
{Vision Transformer with Annotator-Averaged Multi-Label Classification}
{\textcolor{newtext}{This method uses a Vision Transformer (ViT) backbone for frame-level multi-label classification of the three CVS criteria, without temporal modeling or auxiliary objectives.} The ViT encoder, pretrained on ImageNet, extracts spatial features which are passed through a two-layer MLP including ReLU, dropout, and batch normalization. The output is processed via sigmoid activation to produce per-criterion frame-level probabilities.

Training is fully supervised using the binary cross-entropy loss. 
}

\team{Team IRCV-URV}
{EfficientNet-FPN with Attention Mechanisms for
CVS Criteria Classification}
{\textcolor{newtext}{This method uses an EfficientNet-B5 backbone with a Feature Pyramid Network (FPN) and spatial attention for frame-level CVS classification, without temporal modeling or auxiliary objectives.} The architecture is designed to extract multi-scale contextual features relevant to CVS, with the FPN enabling hierarchical feature aggregation and the spatial attention mechanism refining focus on anatomically significant regions implicitly.

Each frame is processed independently and the model is trained using a binary cross-entropy loss with full supervision, using 5-fold cross-validation (80-20 splits), with the final model averaging. 
}

\team{Team mmll}
{Temporal Object-Token Transformer with Graph-Based Ensemble}
{This method ensembles three complementary models for frame-level CVS prediction: a temporal object-token transformer, a non-temporal variant, and a latent graph network. Each model leverages structured object-level representations derived from a Mask R-CNN \citep{he2017mask} detector trained on Endoscapes-Seg50 \citep{murali2023latent}. The detector produces class-labeled bounding boxes, which are used to extract visual features from a separate ResNet-50 backbone. For each object, these features are concatenated with class embeddings and detection scores to form object tokens. To capture spatial relationships, relative layout embeddings are computed between every token pair based on bounding box geometry, and injected as attention biases during transformer encoding.

The temporal model processes object tokens from the current frame and nine preceding frames, supplemented by a global token representing each frame’s overall features. Absolute positional embeddings encode frame order, while spatial biases remain relative. All tokens pass through a 4-layer transformer encoder; the final global token is used for frame-level classification. For the non-temporal model, only the current frame’s objects are used. Each model predicts CVS criteria using MSE loss against the mean annotator label. Final predictions are produced by a weighted ensemble, giving the temporal model twice the weight of each static model.
}

\team{Team Pandas}
{ConvNeXt V2 with Contrastive Learning and Color-Based Augmentation}
{\textcolor{newtext}{This method uses a ConvNeXt V2 backbone for frame-level multi-label classification of the three CVS criteria, with a contrastive auxiliary objective and no temporal modeling.} To enhance feature discrimination, the team incorporates a contrastive learning objective during training, encouraging the model to learn subtle distinctions between frames with different CVS states.

In addition to architectural tuning, the team applies extensive color-based augmentation to improve generalization under diverse imaging conditions. Transformations include color jittering across brightness, contrast, saturation, and hue.
}

\team{Team theator}
{Leveraging visual and temporal transformers for CVS Criteria Analysis}

This method integrates vision-language pretrained transformer backbones with causal temporal modeling to predict CVS criteria and associated uncertainty. The team uses EVA-02-ViT-L, a large-scale Vision Transformer pretrained on ImageNet-1k, as the visual backbone. They train five such models via k-fold cross-validation and one on the full dataset, then aggregate them using the uniform recipe for model soups—an efficient weight-averaging ensembling method that improves generalization under distribution shifts.

The resulting model extracts one feature vector per second for each video, which serves as input to a causal temporal transformer composed of an encoder-decoder architecture. This temporal model predicts frame-level CVS achievement and simultaneously estimates per-frame uncertainty and video-level CVS scores through dedicated heads. The frame model is optimized with focal loss, while the temporal model uses binary cross-entropy for classification and log-cosh loss for uncertainty estimation. Multi-view inference and RandAugment are employed during training to enhance robustness.


\team{Team Transformers}
{Vision Transformer with U-Net-Based Tissue Localization}
{\textcolor{newtext}{This method uses a two-stage segmentation-classification pipeline for frame-level CVS prediction, without temporal modeling.} The first stage uses a U-Net \citep{ronneberger2015u} with an EfficientNet-B0 backbone to segment key anatomical structures involved in the Critical View of Safety. The segmentation model is trained on the Endoscapes-Seg50 dataset to leverage limited annotated surgical data.

The second stage combines the original video frames with the predicted segmentation masks and inputs them into a Vision Transformer (ViT) to classify CVS criteria. This integration allows the classification model to operate with localized anatomical context, potentially improving recognition of subtle cues. 
}

\subsection{Comparative analysis of methods}
\label{sec:theoretical_analysis}

The purpose of this subsection is to summarize and categorize the methodological strategies employed by competing teams in the SAGES CVS Challenge. This categorical grouping provides a structured basis for subsequent performance analysis, allowing trends to be examined both within and across methodological types. Full details per team are given in Table~\ref{tab:method_description}.

\paragraph{Data strategy categories}
\begin{itemize}
    \item \textbf{Standard general pretraining only:} Rely on generic computer vision dataset (e.g., ImageNet) pretraining, without additional surgical datasets.  
    Teams: theator, IRCV-URV, HFUT-MedIA, Caresyntax, CVS HUST, Pandas.
    \item \textbf{Surgical dataset–driven training:} Use models pretrained on surgical datasets before fine-tuning on CVS training data.  
    \begin{itemize}
        \item Detection pretraining on Endoscapes-BBox201: mmll (Mask R-CNN detector), SRV-WEISS (EfficientNet).  
        \item Segmentation pretraining on Endoscapes-Seg50: SRV-WEISS (EndoViT encoder) and SDS-HD (YOLOv8 pseudo-labels)
        \item CVS labels extended from Endoscapes-CVS201: Farm
    \end{itemize}
    \item \textbf{Self-supervised surgical pretraining:}  
    SDS-HD (MoCoV2 on Cholec80, HeiChole, Endoscapes).   
    TUE-VCA (DINO on multi-procedure SurgeNet + GenSurgery datasets). SRV-WEISS (EndoViT initialized with pretrained weights across several distinct surgical procedures \citep{batic2024endovit})

Note that SDS-HD and SRV-WEISS employ both self-supervised pretraining and fully supervised training on additional surgical datasets.

\end{itemize}

\paragraph{Architecture categories}
\begin{itemize}
    \item \textbf{CNN:}  
    FightTumor (ConvNeXt-B),  
    Pandas (ConvNeXt V2),  
    IRCV-URV (EfficientNet-B5 + Feature Pyramid Network with spatial attention),  
    Caresyntax (DenseNet-121).
    
    \item \textbf{Transformer:}  
    Theator \& SDS-HD (EVA-02-ViT-L),  
    HFUT-MedIA (Vision Transformer),  
    TUE-VCA (PVT-v2).
    
    \item \textbf{Hybrid CNN + Transformer:}  
    Transformers (U-Net with EfficientNet-B0 backbone for segmentation + ViT for classification),  
    SRV-WEISS (Dense Prediction Transformer with EndoViT encoder + EfficientNet + ConvNeXt + MLP),  
    mmll (ResNet-50 backbone + Mask R-CNN object detector + Transformer encoders + LG-CVS),  
    Farm (DINOv2, SigLIP, ConvNeXt2-L, InternImage backbones paired with TCN).
\end{itemize}

\paragraph{Temporal modeling}
\begin{itemize}
    \item \textbf{Used temporal modeling:} Theator (temporal transformer), mmll (temporal object-token transformer), Farm (stacked dilated TCNs in some ensemble members), CVS HUST (LSTM).
    \item \textbf{No temporal modeling:} All other teams.
\end{itemize}

\paragraph{Ensembling}
\begin{itemize}
    \item \textbf{Used ensembling:} mmll (3 heterogeneous graph- and transformer-based models), Caresyntax (5 DenseNet-121), Theator (6 EVA-02-ViT-L via model soups), Farm (7 varied backbones), TUE-VCA (5× PVT-v2).
    \item \textbf{No ensembling:} All other teams.
\end{itemize}

\paragraph{Learning objective categories}
\begin{itemize}
    \item \textbf{Classification only:} Caresyntax, FightTumor, HFUT-Media, IRCV-URV, Farm  
    \item \textbf{Classification + auxiliary objectives:} SDS-HD (segmentation loss from YOLOv8 pseudo-labels, label smoothing), SRV-WEISS (segmentation + classification fusion), Transformers (segmentation feeding ViT classification), mmll (graph-based reconstruction loss \citep{murali2023latent}, MSE loss to average rater score), Pandas (contrastive loss), TUE-VCA (L2 reconstruction loss during distillation), Theator (log-cosh loss for uncertainty, BCE for video-level CVS labels, focal loss for backbone), CVS-HUST ($\ell_1$ loss for confidence)
\end{itemize}

Note that SDS-HD, mmll, Theator, CVS-HUST all perform some kind of optimization for uncertainty or confidence estimation.

\section{Results and discussion }

\subsection{Overview}

\paragraph{Headline gains}
Table~\ref{tab:all_subchallenges_one} shows large, measurable improvements over the baseline \textcolor{newtext}{methods} across all three subchallenges. In subchallenge A (mAP), the best score is 69.09 (Farm), a +10.04 absolute gain over \textcolor{newtext}{the highest performing baseline \(LG-DG\)} at 59.05, or a 17.0\% relative improvement; 6 of 13 teams surpass the \textcolor{newtext}{baselines}. In subchallenge B (Brier; lower is better), the best score is 0.022 (theator), \textcolor{newtext}{cutting error by 0.037 points relative to the strongest baseline, ConceptNet, at 0.059, corresponsing to a 62.7\% relative reduction; 9} of 13 teams beat the \textcolor{newtext}{baselines}. In subchallenge C (DRS using the consistent-subset robust-min mAP), the best score is 59.06 (SDS-HD), a +8.44 absolute gain over \textcolor{newtext}{the strongest baseline,} 50.62 for LG-DG, or a 16
.7\% relative improvement; 6 of 13 teams exceed the \textcolor{newtext}{baselines}. Taken together, these headline gains indicate the challenge catalyzed tangible advances over the state of the art.

\paragraph{Common traits of top-performing entries (across subchallenges)}
To provide a compact view of what worked well, we summarize patterns among leaders in each subchallenge using the methodological taxonomy in Section~\ref{sec:theoretical_analysis}:
\begin{itemize}
\item \textbf{Architectures:} Transformer or hybrid transformer–CNN models dominate the very top. Among teams appearing in any top-3 across A/B/C (theator, sds-hd, farm, pandas), 3 of 4 use transformer-based or hybrid designs. In the broader top-5 sets across the three tasks, 7 of 8 unique teams rely on transformers or hybrids.
\item \textbf{Pretraining beyond imagenet:} Leveraging surgical or self-supervised surgical pretraining is common among accuracy/robustness leaders. In both subchallenges A and C, 4 of the top 5 teams use either surgical datasets or self-supervised pretraining on surgical video. Subchallenge B rankings are more mixed on this dimension.
\item \textbf{Ensembling:} Ensembling is frequently present among Subchallenge A 
and C leaders but rare among calibration leaders. In both subchallenge A and C, 4 of the top 5 use ensembles; however, in subchallenge B, only 1 of the top 5 ensembles.
\item \textbf{Temporal modeling:} temporal components appear in 3 of the top 5 teams in subchallenge A and C, despite only 4 of 13 submissions explicitly leveraging temporal context.  
\item \textbf{Learning objectives:} Auxiliary objectives are ubiquitous among top performers. All of the top 5 in subchallenge B employ auxiliary losses or multi-task formulations (e.g., uncertainty, segmentation, contrastive, or distillation objectives). In subchallenges A and C, 4 of the top 5 also incorporate auxiliary objectives.
\end{itemize}

\begin{table*}[t]
\centering
\small
\caption{We show detailed metrics for each subchallenge per CVS criteria (C1--C3) and the average across them. For Subchallenge C (Domain Robustness Score; DRS), we use a \emph{robust-min} evaluation: from the 10 predefined variation splits we drop the worst 10\% (one split) and take the minimum mAP over the remaining splits. We apply this independently to the overall score (computed by first averaging across C1--C3 per sample) and to each criterion (C1--C3). As a result, the average of the per-criterion DRS (C1--C3) may not equal the ``Avg'' DRS, which is obtained by applying the Domain Robustness Score on the average mAP across criteria. The baselines \emph{LG-DG} \textcolor{newtext}{and \emph{ConceptNet} are} included and italicized.}
\label{tab:all_subchallenges_one}
\resizebox{\textwidth}{!}{%
\begin{tabular}{lrrrr lrrrr r lrrrr}
\toprule
\multicolumn{5}{c}{\textbf{Subchallenge A} (mAP, $\uparrow$ is better)} & \multicolumn{5}{c}{\textbf{Subchallenge B} (Brier, $\downarrow$ is better)} &  & \multicolumn{5}{c}{\textbf{Subchallenge C} (DRS, $\uparrow$ is better)}\\
\cmidrule(lr){1-5} \cmidrule(lr){6-10} \cmidrule(lr){12-16}
Team & C1 & C2 & C3 & \textbf{Avg} & Team & C1 & C2 & C3 & \textbf{Avg} &  & Team & C1 & C2 & C3 & \textbf{Avg} \\
\midrule
Farm        & 56.65 & 85.30 & 65.32 & \textbf{69.09} & theator      & 0.024 & 0.023 & 0.020 & \textbf{0.022} &  & SDS-HD      & 41.51 & 73.51 & 41.13 & \textbf{59.06} \\
theator     & 59.06 & 83.61 & 63.94 & \textbf{68.87} & Pandas       & 0.025 & 0.025 & 0.020 & \textbf{0.023} &  & theator     & 39.85 & 76.39 & 46.88 & \textbf{58.11} \\
SDS-HD      & 56.53 & 83.89 & 65.99 & \textbf{68.80} & SDS-HD       & 0.026 & 0.026 & 0.021 & \textbf{0.024} &  & Farm        & 37.08 & 77.13 & 35.35 & \textbf{57.71} \\
mmll        & 54.92 & 78.91 & 58.96 & \textbf{64.26} & SRV-WEISS    & 0.028 & 0.042 & 0.028 & \textbf{0.033} &  & mmll        & 46.37 & 67.64 & 36.37 & \textbf{56.33} \\
TUE-VCA     & 48.12 & 80.60 & 59.94 & \textbf{62.89} & Transformers & 0.032 & 0.050 & 0.031 & \textbf{0.038} &  & TUE-VCA     & 32.71 & 71.61 & 30.57 & \textbf{53.13} \\
Pandas      & 48.24 & 79.82 & 56.42 & \textbf{61.50} & CVS HUST     & 0.038 & 0.055 & 0.036 & \textbf{0.043} &  & Pandas      & 36.86 & 70.39 & 47.74 & \textbf{51.66} \\
\emph{LG-DG (baseline)} & 48.85 & 73.42 & 54.89 & \textbf{59.05} & mmll         & 0.047 & 0.046 & 0.040 & \textbf{0.044} &  & \emph{LG-DG (baseline)} & 40.87 & 60.61 & 37.47 & \textbf{50.62} \\
FightTumor  & 40.85 & 76.34 & 47.15 & \textbf{54.78} & TUE-VCA      & 0.059 & 0.050 & 0.046 & \textbf{0.052} &  & Caresyntax   & 19.72 & 63.79 & 16.82 & \textbf{49.24} \\
Caresyntax  & 40.35 & 75.32 & 44.13 & \textbf{53.27} & Farm         & 0.063 & 0.057 & 0.055 & \textbf{0.058} &  & \emph{\color{newtext}{ConceptNet (baseline)}}     & \color{newtext}{34.23} & \color{newtext}{54.40} & \color{newtext}{23.12} & \textbf{\color{newtext}{44.53}} \\
SRV-WEISS   & 38.16 & 66.27 & 42.29 & \textbf{48.91} & \emph{\color{newtext}{ConceptNet (baseline)}}      & \color{newtext}{0.053} & \color{newtext}{0.065} & \color{newtext}{0.060} & \textbf{\color{newtext}{0.059}} &  & FightTumor  & 27.81 & 63.22 & 30.91 & \textbf{42.83} \\
IRCV-URV    & 34.32 & 69.42 & 37.19 & \textbf{46.98} & Caresyntax   & 0.092 & 0.091 & 0.077 & \textbf{0.086} &  & IRCV-URV    & 26.00 & 59.83 & 14.32 & \textbf{41.38} \\
\emph{\color{newtext}{ConceptNet (baseline)}}     & \color{newtext}{36.03} & \color{newtext}{66.02} & \color{newtext}{37.21} & \textbf{\color{newtext}{46.42}} & IRCV-URV     & 0.102 & 0.108 & 0.094 & \textbf{0.101} &  & SRV-WEISS   & 31.21 & 54.21 & 25.94 & \textbf{40.89} \\
Transformers& 12.98 & 28.94 & 18.17 & \textbf{20.03} & FightTumor   & 0.103 & 0.109 & 0.093 & \textbf{0.102} &  & Transformers& 11.39 & 23.50 &  6.44 & \textbf{18.01} \\
CVS HUST    & 13.76 & 25.14 &  9.15 & \textbf{16.01} & \emph{LG-DG (baseline)} & 0.130 & 0.119 & 0.121 & \textbf{0.124} &  & HUFT-MedIA  &  8.01 & 21.94 &  7.80 & \textbf{13.65} \\
HUFT-MedIA  &  9.91 & 23.44 &  9.51 & \textbf{14.29} & HUFT-MedIA   & 0.119 & 0.172 & 0.110 & \textbf{0.134} &  & CVS HUST    &  8.11 & 19.58 &  5.69 & \textbf{11.65} \\
\bottomrule
\end{tabular}}
\end{table*}

\subsection{Cross–subchallenge relationships}
\label{sec:cross_subchallenge}

\begin{table}[ht]
\centering
\caption{Ranks for Subchallenge A (CVS Classification), Subchallenge B (Uncertainty Quantification), and Subchallenge C (Robustness) are shown. A lower rank number indicates a better performance.}
\label{tab:rank_stability}
\resizebox{0.5\textwidth}{!}{%
\begin{tabular}{lcccc}
\toprule
\textbf{Team} & \textbf{Overall Rank} & \textbf{Subchallenge A} & \textbf{Subchallenge B} & \textbf{Subchallenge C} \\
\midrule
theator & 1 & 2 & 1 & 2 \\
SDS-HD & 2 & 3 & 3 & 1 \\
Farm & 3 & 1 & 9 & 3 \\
Pandas & 4 & 6 & 2 & 6 \\
mmll & 5 & 4 & 7 & 4 \\
TUE-VCA & 6 & 5 & 8 & 5 \\
SRV-WEISS & 7 & 9 & 4 & 10 \\
Caresyntax & 8 & 8 & 10 & 7 \\
FightTumor & 9 & 7 & 12 & 8 \\
Transformers & 10 & 11 & 5 & 11 \\
IRCV-URV & 11 & 10 & 11 & 9 \\
CVS\_HUST & 12 & 12 & 6 & 13 \\
HFUT-MedlA & 13 & 13 & 13 & 12 \\
\bottomrule
\end{tabular}%
}
\end{table}

Using the ranks in Table~\ref{tab:rank_stability}, A and C track each other closely (Spearman = 0.96), while A vs B and B vs C are only weakly aligned (both 0.38). This is consistent with Table~\ref{tab:all_subchallenges_one}: teams that classify well (A) also tend to be robust (C), whereas B behaves more independently.

Method traits by top five
counting presence within the top five of each subchallenge:
\begin{itemize}
\item \textbf{Pretraining beyond ImageNet:} A 4/5, C 4/5, B 2/5
\item \textbf{Ensembling:} A 4/5, C 4/5, B 1/5
\item \textbf{Temporal modeling:} A 3/5, C 3/5, B 1/5
\end{itemize}
in short, the ingredients that lead to robust performance are not prerequisites for capturing clinical ambiguity.

\subsection{Focused Analysis: Overall Performance}
\label{sec:focused_overall}

Table~\ref{tab:expert_bounds_vs_top5_full} shows clear criterion-wise differences within Subchallenge A. Macro-F1 is consistently highest on C2 and lowest on C1 (with C3 in between), while per-criterion accuracies remain uniformly high (about 86–92 percent). This gap between accuracy and macro-F1 is consistent with the fact that most frames capture negative examples of the CVS criteria: teams (and the mode-of-raters labels) agree on negatives, but miss more positives—macro-F1 exposes those misses in a way accuracy does not.

To anchor these results, we compare against two expert references, evaluated against the same mode-of-raters labels. Expert (upper bound) predicts the per-frame, per-criterion mode label and thus attains 100 percent by construction. Expert (lower bound) predicts the consensus on unanimous frames and the minority label on frames where there was partial agreement between the 3 annotators.  These robust expert references are based on 19 annotators who underwent the training and quality–control pipeline described in Section~\ref{sec:endoglacier} and, within the 300–video test set, each annotated 47.4±25.0 videos on average. The top five Subchallenge A teams in Table~\ref{tab:expert_bounds_vs_top5_full} exceed this floor by an average of about 23.4 macro-F1 points (60.30 vs 36.89) and 19.0 accuracy points (75.99 vs 56.96), indicating performance that these systems are approaching expert-level capability at CVS assessment. 

\begin{table}[t]
\centering
\small
\caption{Overall and per-criterion macro-F1 and accuracy for the top five Subchallenge~A teams (ordered by Subchallenge~A mAP in Table~\ref{tab:all_subchallenges_one}), compared to expert references. Accuracy is reported as overall subset accuracy (all three criteria correct) and per-criterion accuracy; macro-F1 is reported overall and per criterion. Expert (upper bound) predicts the per-frame, per-criterion rater-mode label; Expert (lower bound) predicts the consensus on unanimous frames and the minority label on disagreement frames. All values are percentages.}
\label{tab:expert_bounds_vs_top5_full}
\resizebox{\linewidth}{!}{%
\begin{tabular}{lrrrrrrrr}
\toprule
 & \multicolumn{4}{c}{macro-F1} & \multicolumn{4}{c}{accuracy} \\
\cmidrule(lr){2-5}\cmidrule(lr){6-9}
Team & Overall & C1 & C2 & C3 & Overall & C1 & C2 & C3 \\
\midrule
Expert (upper bound) & 100.00 & 100.00 & 100.00 & 100.00 & 100.00 & 100.00 & 100.00 & 100.00 \\
Farm                 & 62.94 & 53.64 & 76.09 & 59.10 & 77.65 & 89.85 & 89.35 & 92.26 \\
theator              & 64.30 & 55.34 & 75.13 & 62.42 & 75.59 & 89.93 & 87.59 & 92.24 \\
SDS-HD               & 61.12 & 51.31 & 74.41 & 57.64 & 77.33 & 90.37 & 88.33 & 92.41 \\
mmll                 & 59.35 & 52.20 & 71.54 & 54.30 & 73.11 & 88.74 & 86.31 & 90.06 \\
TUE-VCA              & 53.81 & 39.05 & 69.76 & 52.63 & 76.28 & 89.54 & 87.20 & 91.83 \\
Expert (lower bound) & 36.89 & 31.12 & 47.44 & 32.11 & 56.96 & 77.30 & 72.83 & 82.15 \\
\bottomrule
\end{tabular}}%
\end{table}

\begin{figure*}[ht]
         \centering
\includegraphics[width=\linewidth]{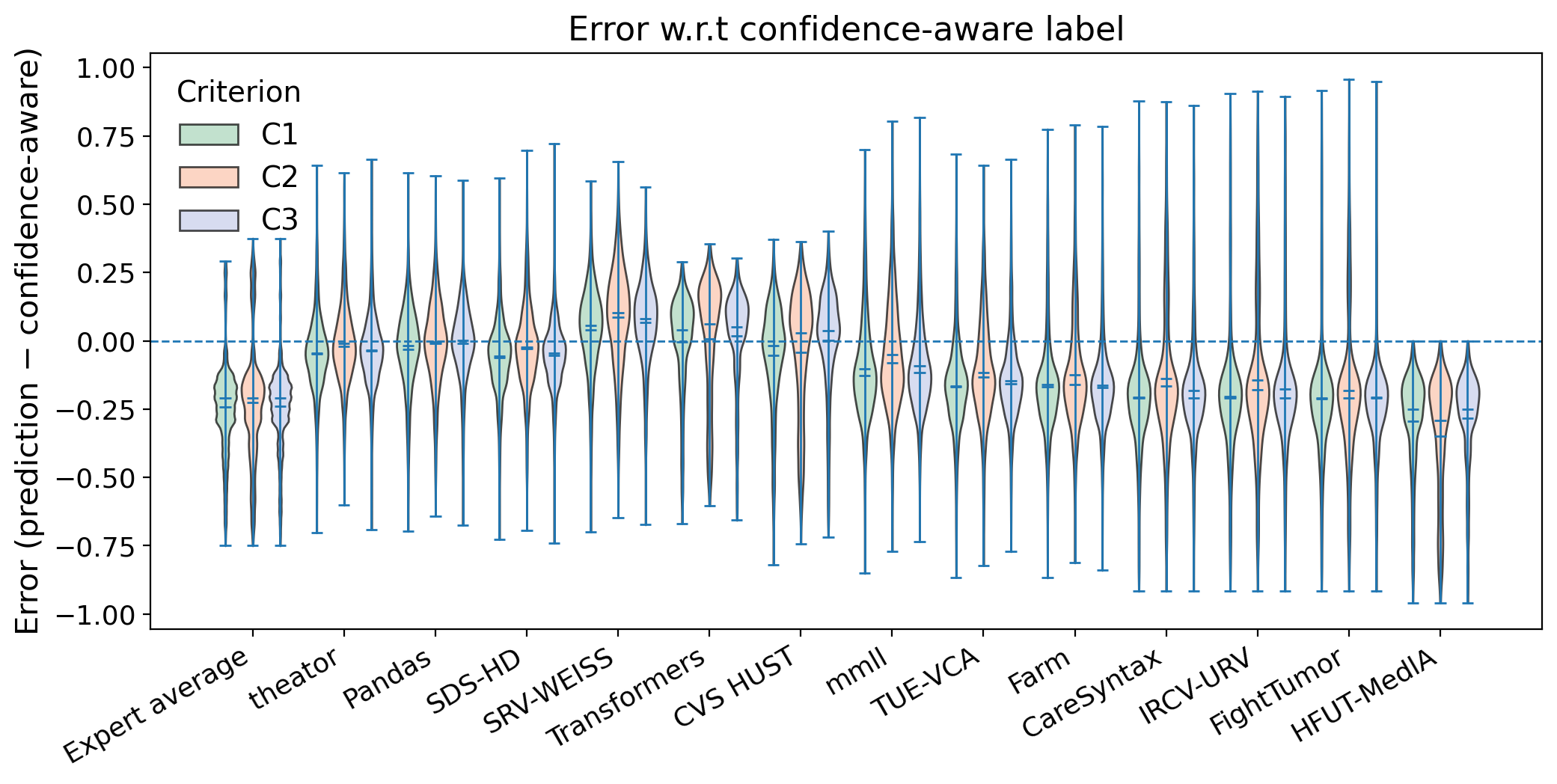} 
    \caption{Violin plots of signed error relative to the confidence-aware label described in Section~\ref{sec:challenge}. Each group shows one team, with the three CVS criteria (C1, C2, C3) in different colors. Error is defined as predicted probability minus the confidence-aware target for a criterion; zero (dashed line) means perfect alignment with the target, while larger absolute values indicate greater deviation. Positive values indicate a tilt toward the positive label relative to the target; negative values indicate a tilt toward the negative label. The Expert average violin on the left shows the difference between the mean rater label and the confidence-aware label, providing a human reference. Teams are displayed left to right in order of Subchallenge~B performance.}
    \label{fig:violin_by_criterion}
\end{figure*}

\subsection{Focused Analysis: Calibration (Subchallenge B)}
\label{sec:focused_calibration}

Subchallenge~B evaluates how well a model's predicted probabilities align with a confidence-aware target (Section~\ref{sec:challenge}): the target approaches 0.5 when raters are uncertain and moves toward 0 or 1 when they are confident. \textcolor{newtext}{As a result, models are encouraged not only to be accurate, but also to express uncertainty in regions where expert agreement is low or confidence is limited. }Figure~\ref{fig:violin_by_criterion} visualizes, per team and per criterion (C1--C3), the signed error $p - y_{\text{conf}}$ (predicted probability minus confidence-aware label). Distributions centered tightly around 0 indicate good calibration; large absolute deviations and wide spreads indicate miscalibration.

\paragraph{Who is calibrated? Rankings mirror the violins}
Top Subchallenge~B teams theator, Pandas, and SDS-HD show small biases and relatively narrow spreads across criteria (means within about $\pm 0.06$).

\paragraph{Auxiliary objectives, uncertainty objectives, and localization}
All 7 top-ranking teams in subchallenge B, without exception, make use of auxiliary objectives. Among them, the top performers (theator, Pandas, SDS-HD) achieve tight, near-zero distributions, suggesting auxiliary supervision can help calibration when paired with strong training recipes. Explicit uncertainty objectives (SDS-HD, theator, mmll, CVS HUST) are not a guarantee on their own: theator and SDS-HD are well-centered and near the top; CVS HUST is mixed; mmll remains negatively biased. Similarly, using localization signals (SDS-HD, SRV-WEISS, Transformers, mmll) yields heterogeneous outcomes: excellent for SDS-HD, positively tilted for SRV-WEISS, broader for Transformers, and under-confident (negative) for mmll. In short, how these ingredients are integrated matters as much as their presence.

\paragraph{Direction of error and what models learn}
Relative to the confidence-aware target, most mid- and lower-ranked systems skew negative (probabilities below $y_{\text{conf}}$), indicating conservative estimates when raters express partial confidence. Top teams nevertheless sit much closer to $y_{\text{conf}}$ than to the hard mean label, indicating they learn to place probabilities in the soft region that reflects rater uncertainty, whether they learn to do so implicitly or explicitly.

\paragraph{A note on Farm}
Farm is typically top-3 in Subchallenges~A/C, yet shows sizable negative bias and wider spread in Subchallenge~B, aligning with its drop in Brier ranking in Table~\ref{tab:all_subchallenges_one}. This supports the takeaway that optimizing for discriminative performance (for example, mAP) does not inherently yield confidence alignment; calibration requires targeted choices.

\begin{figure}[ht]
         \centering
\includegraphics[width=\linewidth]{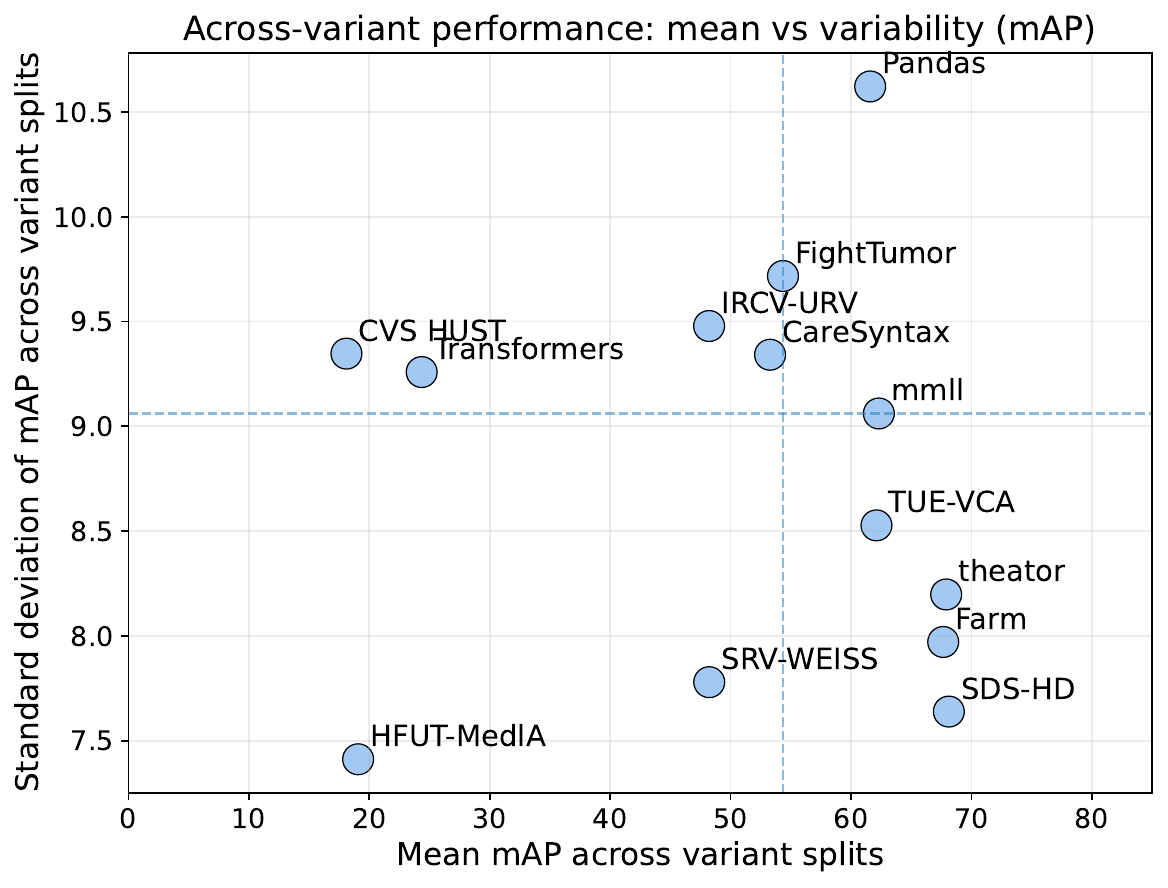} 
    \caption{Mean mAP (x-axis) vs. standard deviation across variant splits (y-axis) for each team. Toward the right indicates better average performance; toward the bottom indicates more consistent performance across splits.}
    \label{fig:mean-std-variant-splits}
\end{figure}

\subsection{Focused Analysis: Robustness across Variant Splits}
Figure~\ref{fig:mean-std-variant-splits} summarizes each team’s behavior by plotting mean mAP across the variant test splits of Subchallenge C against its standard deviation. We did not observe statistically significant methodological patterns, within the submitted methods, linking specific architectural or training choices to robustness. Instead, teams populate all four quadrants of the plot: some achieve high means with low dispersion, others reach similar means with notably higher dispersion, and several exhibit uniformly lower means with either narrow or wide spread. This heterogeneity reinforces a practical takeaway for deployment: optimizing average performance alone is not sufficient. Methods that explicitly prioritize, or at minimum validate, equitable performance across different data distributions are needed to ensure responsible and fair real-world deployment.

\section{Limitations \textcolor{newtext}{and Outlook}}
\textcolor{newtext}{Our work comes with several limitations that we would like to be transparent about, while also outlining directions for future progress.}

\textcolor{newtext}{A limitation of challenge design in general, including ours, is that it prioritises broad exploration of design choices over definitive conclusions about any single choice. In return, we surfaced clear, community-relevant signals across diverse architectures, auxiliary objectives, and training strategies. Rather than providing final answers, the benchmark establishes a strong empirical foundation and a shared reference point for focused, controlled ablations and methodological refinement. In parallel, the dataset and associated annotations enable a range of more targeted investigations, including stratified analyses across cohorts, identification of failure modes, and data-centric studies aimed at improving clinical reliability.}

\textcolor{newtext}{A second limitation arises from practical trade-offs required to operate under a fixed, albeit substantial, annotation budget. To maximise diversity across institutions, surgical practices, and case complexity, we prioritised breadth over density. This led to several design choices: restricting annotations to a 90-second decision-critical window, sampling frames sparsely within that window, and capturing annotator confidence at the video level rather than per frame. While these decisions enabled large-scale, globally representative data collection from a highly constrained expert workforce, they may omit finer temporal dynamics, rare events, or frame-level variations in uncertainty. Future work may explore alternative, or supplementary, allocation strategies, including denser temporal annotation, extended procedural coverage, and finer-grained uncertainty modeling.}

\textcolor{newtext}{More broadly, the benchmark is intentionally scoped to a clinically grounded computer vision task: frame-level assessment of CVS criteria within a defined temporal window. While this task is relevant to a wide range of downstream applications, including training and credentialing, retrospective audit, real-time intraoperative guidance, and medico-legal or reimbursement contexts, each of these applications introduces distinct requirements that are not fully captured by the current benchmark. These include how predictions should be aggregated over time to produce video- or procedure-level assessments, how temporal stability and persistence should be enforced, how models should operate under compute and latency constraints, and how predictions should be presented in an interpretable and clinically actionable manner. In high-risk or regulatory settings, additional considerations may include performance on rare or difficult cases, validation on independent and geographically diverse datasets, and the incorporation of more interpretable modeling components, such as anatomically grounded representations. Addressing these dimensions will require dedicated evaluation protocols and, in some cases, rethinking both modeling and annotation strategies.}

\textcolor{newtext}{Finally, while the EndoGlacier framework enabled the orchestration of large-scale data collection and annotation across multiple institutions, the overall effort, spanning approximately three years, highlights the difficulty of scaling such initiatives. This work provides immediate value in several ways: it establishes a large, diverse benchmark dataset; offers a snapshot of the state of the field across multiple modeling paradigms; and introduces a modular blueprint for coordinating complex, multi-actor pipelines. At the same time, parts of the system rely on proprietary or custom-built components, and significant effort was required for coordination, governance, and quality control. We designed the pipeline to be modular and explicitly documented its components and availability, with the expectation that future iterations will benefit from increased standardization, more accessible tooling, and broader institutional support. This may include greater involvement of surgical societies, shared infrastructure across research groups, and a gradual shift toward more open and interoperable ecosystems for surgical data science.}

\section{Conclusion}

The SAGES Critical View of Safety (CVS) Challenge brought together clinicians and engineers, industry and academia, in a coordinated effort to address a single, clinically relevant problem: the automated assessment of a key surgical safety step. CVS serves as an exemplar of surgical quality assessment, clear in its definition, supported by strong clinical evidence, yet inconsistently performed worldwide. The resulting benchmark spans continents, involves data from 54 hospitals, and engaged hundreds of clinicians in its creation, offering a rare real-world testbed for AI-based surgical quality assessment.

Delivering this benchmark required infrastructure capable of sustaining global collaboration at scale. The EndoGlacier framework enabled the orchestration of diverse video and annotation flows, rigorous multi-annotator quality control, and the coordination of contributors across institutions and countries. We hope its design will serve as \textcolor{newtext1}{a transferable blueprint} for future large-scale benchmarks in surgery.

The challenge attracted submissions from 13 international teams and produced substantial advances over the baseline across all evaluation axes. In Subchallenge~A, the best method improved mean average precision by 17\% relative; in Subchallenge~B, the top performer reduced Brier error by over \textcolor{newtext}{62}\%; and in Subchallenge~C, the leading method achieved a 17\% relative gain in domain robustness. These results underscore both the potential of current methods and the importance of evaluating accuracy, calibration, and robustness together when developing clinically deployable systems.

Methodological trends observed across top-performing submissions, including the effectiveness of transformer-based or hybrid architectures, surgical or self-supervised surgical pretraining, temporal modeling, and auxiliary learning objectives, provide a springboard for future research. As such, this benchmark not only measures progress but also guides it, helping the field move toward robust, trustworthy AI systems capable of improving surgical safety and quality worldwide.

\section*{Acknowledgment }
{\small
We would like to thank Surgical Safety Technologies for their support in enabling the video collection and de-identification infrastructure. We would like to thank all contributing members to each of the submissions, including 
Leon Mayer, Patrick Godau, Piotr Kalinowski, Lars Kr\"{a}mer, \"{O}mer S\"{u}mer, Dominik Michael, Tim R\"{a}dsch, Fabian Isensee, Lena Maier-Hein, 
Sanket Rajan Gupte, Elaine Sui, Alan Brown, Josiah Aklilu, Reid Dale, Charlotte Egeland, Jeffrey Heo, Matthew Leipzip, Anita Rau, Joshua Villarreal, Jeffrey Jopling, Dan Azagury, Serena Yeung-Levy, 
Pranav Poudel, Sachin Acharya, 
Sophia Bano, Euan Sexton, 
Yuanbin Wang, 
Dom\'enec Puig, 
Chengzhi Hu, Zhiwei Wang, 
Yuxuan Yang, Rui Xu, Shuai Ding.
The SAGES CVS Challenge Data Consortium is represented by Brian Quaranto, Mika Makari, Omero Costa Filho, Giovanni Taffurelli, Alexander Farrell, Taner Shakir, Henry Badgery, Alexandros Chamzin, Fernando Ponce Leon, Salvador Morales-Conde, Lovenish Bains, Benjamin Babic, Micheli Domingos, Yuri Fishman, Ankit Patel, Ahmed Abdo, Antonio S Soares, Brian Davis, Elsie Vazquez, Garrett Skinner, Gourav Mittal, Ionela Daniela Gal, Jan Liechti, Jo{\"e}l Lavanchy, Kohei Mishima, Lachlan Dick, Luca Gordini, Maria Vannucci, Michelle Liu, Nadine Muensterer, Panagiotis Kapsampelis, Paul Wisniowski, Riya Manas Sharma, Yaacov Davidow, Yunfan Li.
}

\section*{Funding }
{\small

This work was supported by French state funds managed by the ANR under Grant ANR-20-CHIA-0029-01 (Chair AI4ORSafety) and Grant ANR-10-IAHU-02 (IHU Strasbourg), and by BPI France under reference DOS0180017/00 (project 5G-OR). This work was also developed within the Interdisciplinary Thematic Institute HealthTech (ITI 2021-2028 program of the University of Strasbourg, CNRS and Inserm) supported by the IdEx Unistra (ANR-10-IDEX-0002) and SFRI (STRATUS project, ANR-20-SFRI-0012)\textcolor{newtext}{, and within the AI Cluster Grand Est ENACT (ANR-23-IACL-0004)}. This work has received funding from the European Union (ERC, CompSURG, 101088553). Views and opinions expressed are however those of the authors only and do not necessarily reflect those of the European Union or the European Research Council. Neither the European Union nor the granting authority can be held responsible for them. This work was also partially supported by a CRICO grant. Awards for the challenge were sponsored by NVIDIA, Olympus, Intuitive and Medtronic plc.


Participating teams would like to acknowledge the following funding: 
%
%
{\it\bf Farm}: Wellcome Leap
%
%
{\it\bf SDS-HD}: National Center for Tumor Diseases (NCT) Heidelberg, Helmholtz Imaging
{\it\bf SRV-WEISS}: EPSRC (EP/Y01958X/1,EP/W00805X/1)
%
%
%
%
%
%
%
%
}

\section*{Declaration of generative AI and AI-assisted technologies in the writing process}
{\small
During the preparation of this work, the authors used ChatGPT (OpenAI) to refine language, improve readability, and ensure clarity of expression. After using this tool, the authors reviewed and edited the content as needed and take full responsibility for the final version of the manuscript.
}

\section*{Data and code availability}
{\small
\textcolor{newtext}{The dataset and benchmark evaluation code for metric calculation is publicly accessible within the challenge repository\footnote{\url{https://huggingface.co/datasets/CAMMA-public/SAGES_CVS_Challenge_2024}}}.
}

\section*{CRediT authorship contribution statement}
{\small
{\bf DA} : Conceptualization, Data Curation, Data Analysis and Interpretation, Methodology, Software, Investigation, Validation, Evaluation, Formal Analysis, Visualization, Writing - Original Draft, Writing - Review \& Editing, Challenge Organization, Resources.
{\bf JE} : Conceptualization, Data Curation, Data Analysis and Interpretation, Methodology, Writing - Review \& Editing, Challenge Organization, Resources.
{\bf ZL, YB, JPM, SC} : Data Curation, Data Analysis and Interpretation, Methodology, Writing - Review \& Editing, Challenge Organization.
{\bf KY} : Data Analysis and Interpretation, Methodology, Writing - Review \& Editing.
{\bf PHC, NZ, SC} : Data Curation, Methodology, Writing - Review \& Editing.
{\bf DN, OB, AY, JH, XW, RL, LL, YW, SK, BB, TJ, ZM, AW, JM, YX, ZW, AMOA, HAFR, BS, KY, YZ, HW} : Methodology, Software, Writing - Review \& Editing.
{\bf QL, FF, XL, PM, DAH, GR, OM, NP} : Conceptualization, Writing - Review \& Editing, Supervision, Challenge Organization, Resources, Funding Acquisition, Project Administration.
}
\ifdefined\arxiv
\section*{CRediT authorship contribution statement}
{\small
{\bf DA} : Conceptualization, Data Curation, Data Analysis and Interpretation, Methodology, Software, Investigation, Validation, Evaluation, Formal Analysis, Visualization, Writing - Original Draft, Writing - Review \& Editing, Challenge Organization, Resources.
{\bf JE} : Conceptualization, Data Curation, Data Analysis and Interpretation, Methodology, Writing - Review \& Editing, Challenge Organization, Resources.
{\bf ZL, YB, JPM, SC} : Data Curation, Data Analysis and Interpretation, Methodology, Writing - Review \& Editing, Challenge Organization.
{\bf KY} : Data Analysis and Interpretation, Methodology, Writing - Review \& Editing.
{\bf PHC, NZ, SC} : Data Curation, Methodology, Writing - Review \& Editing.
{\bf DN, OB, AY, JH, XW, RL, LL, YW, SK, BB, TJ, ZM, AW, JM, YX, ZW, AMOA, HAFR, BS, KY, YZ, HW} : Methodology, Software, Writing - Review \& Editing.
{\bf QL, FF, XL, PM, DAH, GR, OM, NP} : Conceptualization, Writing - Review \& Editing, Supervision, Challenge Organization, Resources, Funding Acquisition, Project Administration.
}
\fi

\bibliographystyle{model2-names.bst}\biboptions{authoryear}
\bibliography{main}
%

\end{document}